\documentclass{article}

\usepackage{microtype}
\usepackage{graphicx}
\usepackage{booktabs} % for professional tables

\usepackage{algorithm}
\usepackage{algorithmic}
\usepackage{wrapfig}
\usepackage{amsmath}
\usepackage{amssymb}
\usepackage{mathtools}
\usepackage{amsthm}
\usepackage{fontawesome5}
\usepackage[textsize=tiny]{todonotes}
\definecolor{uclablue}{rgb}{0.15, 0.45, 0.68}
\usepackage[
    pagebackref,
    breaklinks,
    colorlinks=true,
    citecolor=uclablue,
    linkcolor=uclablue,
    urlcolor=uclablue
]{hyperref}
\usepackage[capitalize,noabbrev]{cleveref}
\usepackage[preprint]{neurips_2026}

\theoremstyle{plain}

\theoremstyle{definition}

\theoremstyle{remark}

\title{Learning from Trials and Errors: Reflective Test-Time Planning for Embodied LLMs}

\author{%
\small
Yining Hong$^{1}$,
Huang Huang$^{1}$,
Manling Li$^{2}$,
Li Fei-Fei$^{1}$,
Leonidas Guibas$^{1}$,
Jiajun Wu$^{1}$,
Yejin Choi$^{1}$\\[3pt]
$^{1}$Stanford University \qquad
$^{2}$Northwestern University
}

\begin{document}

\maketitle
\vspace{-3em}
\begin{center}
\small
\faGlobe\ Website: \href{https://reflective-test-time-planning.github.io}{https://reflective-test-time-planning.github.io} \\
\faGithub\ Code: \href{https://github.com/Reflective-Test-Time-Planning/Reflective-Test-Time-Planning}{https://github.com/Reflective-Test-Time-Planning/Reflective-Test-Time-Planning} \quad
\end{center}
\maketitle
% \vspace{-2em}
\begin{figure}[htbp]
  \centering
  \includegraphics[width=\textwidth]{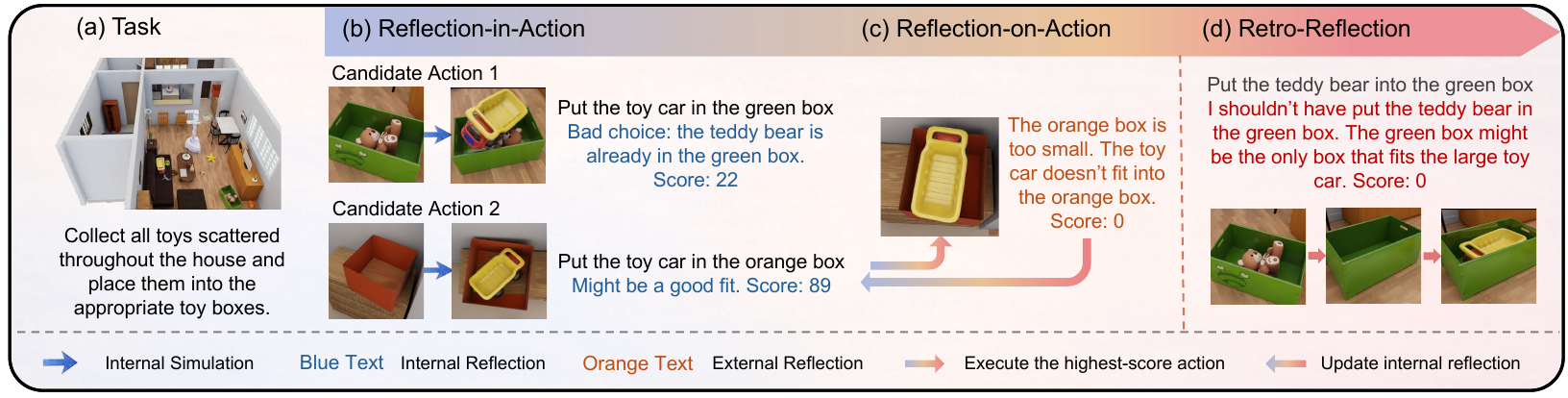}
  \vspace{-1.5em}
  \caption{Conceptual overview of Reflective Test-Time Planning. The agent (a) receives a long-horizon task,
  (b) performs \textit{reflection-in-action} by internally simulating and scoring candidate actions,
  (c) performs \textit{reflection-on-action} by updating its beliefs and decision systems based on execution outcomes,
  and (d) conducts \textit{retrospective reflection} to revise earlier decisions with hindsight.}
  \label{fig:teaser}
\end{figure}

\begin{abstract}
\vspace{-0.5em}
Embodied LLMs endow robots with high-level task reasoning, but they cannot reflect on what went wrong or why, turning deployment into a sequence of independent trials where mistakes repeat rather than accumulate into experience. Drawing upon human reflective practitioners, we introduce Reflective Test-Time Planning, which integrates two modes of reflection: \textit{reflection-in-action}, where the agent uses test-time scaling to generate and score multiple candidate actions using internal reflections before execution; and \textit{reflection-on-action}, which uses test-time training to update both its internal reflection model and its action policy based on external reflections after execution. We also include retrospective reflection, allowing the agent to re-evaluate earlier decisions and perform model updates with hindsight for proper long-horizon credit assignment. Experiments on our newly-designed Long-Horizon Household benchmark and MuJoCo Cupboard Fitting benchmark show significant gains over baseline models, with zero-shot generalization to photorealistic HM3D environments and real-robot experiments on a Franka Panda arm. Ablations confirm that reflection-in-action and reflection-on-action are mutually dependent, and that retrospective reflection achieves better credit assignment than step-wise external feedback at lower computational overhead. Qualitative analyses further highlight behavioral correction through reflection.
\end{abstract}

\vspace{-0.5em}
\begin{quote}
\textit{``Error isn't simple darkness, it sheds a light of its own.''}
\vspace{-0.5em}
\begin{flushright}
--- Kathryn Schulz, \textit{Being Wrong}
\end{flushright}
\end{quote}

\vspace{-0.5em}

\section{Introduction}
\vspace{-0.5em}
% Embodied large language models \cite{brohan2023rt2, kim2024openvla, hong2024multiply} have unlocked new possibilities for robotic reasoning, enabling agents to understand complex instructions and decompose long-horizon tasks into actionable plans. Yet beneath this sophisticated semantic capability lies a fundamental brittleness: these systems operate as static oracles, dispensing decisions based solely on patterns absorbed during pretraining. When actions fail, they cannot reflect on what went wrong or what caused the failure. Thus, deployment becomes a sequence of independent trials where the same mistakes repeat rather than an accumulation of experience.

Embodied LLMs \citep{brohan2023rt2, kim2024openvla, hong2024multiply} equip agents with task planning abilities, but they remain brittle static oracles that cannot learn from failures, turning deployment into independent trials of repeated mistakes rather than accumulated experience. Humans, in contrast, are natural reflective practitioners. Drawing on Schön's framework about reflective planning \citep{Schon1983ReflectivePractitioner}, humans fluidly alternate between two modes of reflection: through \textit{reflection-in-action}, we engage in internal simulation, questioning whether our planned approach will actually work given what we currently understand; through \textit{reflection-on-action}, we use the actual outcomes to reshape both our beliefs about the environment and our strategies for acting within it. An illustrative example of these reflection modes is shown in Figure~\ref{fig:teaser}. This bidirectional flow allows us to learn not only from outcomes, but also from the very process of engaging with an uncertain world.

Current approaches have pushed forward individual reflection mechanisms, but largely in isolation. One line of work \citep{shinn2023reflexion, madaan2023selfrefine} uses LLM-based verbal reflection, generating natural-language critiques of past behavior to condition future actions. While this enables reflection-on-action at the level of reasoning traces, it remains underexplored how such reflections can move beyond contextual text to become persistent learning signals. A second line of work \citep{zhen20243dvla, feng2025reflectiveplanningvisionlanguagemodels} supports reflection-in-action by guiding action selection through internal world models, yet raising another challenge: can such internal beliefs be updated when execution reveals mismatches between expected and actual outcomes? Classical planning and RL methods \citep{rubinstein1999cem, williams2017mppi, schulman2017ppo} approach the problem from a different angle, revising behavior through scalar reward signals aggregated over many rollouts, episodes, or resets. Yet such scalar feedback is non-linguistic, noisy, and amortized across episodes, making it hard to diagnose repeated failures within just one single unfolding episode of interaction.

To address these challenges and operationalize both reflection modes in embodied settings, we introduce Reflective Test-Time Planning, a systematic framework that unifies reflection-in-action and reflection-on-action for embodied agents during test-time deployment. Concretely, the framework employs three embodied LLMs during deployment: an action generation model $\pi_\theta$, an internal evaluator $V_{\phi_i}$, and an external evaluator $V_{\phi_e}$. During reflection-in-action, the agent samples $N$ candidate actions via high-temperature sampling, uses $V_{\phi_i}$ to generate internal reflections scoring each candidate, then executes the highest-scoring action. After execution, $V_{\phi_e}$ generates an external reflection, providing an immediate, language-based evaluation of what happened and why.

This immediate external reflection grounds beliefs in reality, but remains inherently local—it only evaluates consequences visible at the next timestep. Many embodied failures are non-local: an action that appears successful may later block progress, and a seemingly suboptimal action may enable future success. To address this temporal credit assignment problem, we introduce retrospective reflection, where $V_{\phi_e}$ periodically re-evaluates earlier decisions with hindsight (e.g., at room transitions or after repeated failures). These hindsight assessments provide self-supervised signals at deployment time, enabling two forms of test-time training: (1) policy gradient for $\pi_\theta$ to favor actions that score well under hindsight, and (2) supervised learning for $V_{\phi_i}$ to anticipate what hindsight will reveal. Because these updates revise not only the action policy but also the predictive assumptions behind it, the process constitutes a form of double-loop learning \citep{Argyris1977DoubleLoop}, in which agents learn not merely from outcomes but from diagnosing and correcting the underlying causes of their errors.

We evaluate our approach on two embodied benchmarks that we design to stress error-driven adaptation: (1) a Long-Horizon Household benchmark that requires failure recovery during multi-step planning across rooms, and (2) a controlled MuJoCo Cupboard Fitting benchmark that isolates geometric placement failures. Our framework achieves large gains over reflective language, RL and world-model baselines, with zero-shot generalization to photorealistic HM3D environments and real-robot experiments on a Franka Panda arm. Ablations indicate that improvement emerges only when both reflection-in-action and reflection-on-action take place, and when both action policy and internal reflection model are updated during deployment; and that retrospective reflection achieves better credit assignment than step-wise external feedback at lower computational overhead. Qualitative analyses further show that reflection reduces repetitive failure modes in practice.

% To summarize, our contributions are:
% \begin{itemize}
% \item We propose Reflective Test-Time Planning, a framework that unifies reflection-in-action (prospective internal simulation) and reflection-on-action (retrospective external critique) for embodied agents.
% \item We introduce a verbal reflection pipeline consisting of internal, external, and retrospective reflections, and convert these signals into test-time scaling and test-time training updates for both the action policy and internal predictive model.
% \item We design two embodied benchmarks for error-driven adaptation: a BEHAVIOR-style long-horizon household suite and a MuJoCo Cupboard Fitting benchmark.
% \item We demonstrate substantial gains over reflective LLM baselines and world-model-based approaches, with ablations highlighting the complementarity of reflection-in-action and reflection-on-action. Qualitative analyses, including preliminary real-robot deployments, further show how reflection enables agents to diagnose failures and recover from them.
% \end{itemize}
\vspace{-0.5em}
\section{Related Works}
\vspace{-1em}

\textbf{Test-Time Adaptation (TTA).}
TTA adapts models to distribution shifts during inference without source data \citep{sun2020ttt, wang2021tent, liang2023comprehensive}. Early methods minimize entropy, with Tent \citep{wang2021tent} updating batch-norm parameters online and later work adding calibrated objectives \citep{niu2022efficient, tan2024ceta}. Parameter-efficient TTA uses LoRA \citep{hu2021lora, liu2025lora}, bias-only tuning \citep{dumpala2023testtimetrainingspeech}, or hidden-state adaptation for long-context memory \citep{sun2024ttt}. In embodied settings, continual learning supports manipulation and navigation \citep{lesort2020continual, meng2025legion, hajizada2024continual, mendonca2024continuous}. We adapt models at test time via self-supervised signals from the agent's own verbal assessments.

\textbf{Multimodal Embodied Large Language Models.}
Recent foundation models leverage large-scale robotic data for zero-shot generalization \citep{brohan2023rt2, kim2024openvla, driess2023palme}, including RT-2's web-knowledge transfer and OpenVLA's heterogeneous embodiment support. 3D spatial grounding spans point clouds \citep{hong20233dllm}, 3D patches \citep{zhu2024llava3d}, and lightweight point-cloud injection \citep{xu2025pointvla}. Extensions incorporate multisensory interaction \citep{hong2024multiply}, generative world models \citep{zhen20243dvla}, long-term spatial--temporal memory \citep{hu20253dllmmemlongtermspatialtemporalmemory}, interleaved multimodal instructions \citep{chen2025interleave}, and chain-of-thought reasoning \citep{zhao2025cotvla, mu2023embodiedgptvisionlanguagepretrainingembodied}. In contrast, we treat deployment as a learning phase in which the embodied multimodal LLM reflects on actions and updates itself via test-time training.

\textbf{Reflection and Self-Improvement in AI Agents.}
Verbal self-reflection methods such as Reflexion \citep{shinn2023reflexion} store natural-language critiques to guide future actions, with extensions to self-refinement \citep{madaan2023selfrefine, chen2024selfcontrast}, tool-assisted verification \citep{gou2024critic}, curiosity-driven reflection \citep{kauvar2024curious}, multi-agent systems \citep{ng2024reflection}, and robotics \citep{huang2022inner}. These methods support reflection-on-action but use reflections as text rather than deployment-time parameter updates, while complementary reflection-in-action methods use internal predictive models to anticipate outcomes \citep{zhen20243dvla, feng2025reflectiveplanningvisionlanguagemodels, zhen2025learning4d, hafner2024masteringdiversedomainsworld}. Our method unifies both modes by converting reflections into self-supervised training signals for parameter updates during deployment.

\textbf{Planning and Reinforcement Learning for Embodied Agents.}
Our work relates to, but differs from, classical planning and RL. Planning methods such as CEM \citep{rubinstein1999cem} and MPPI \citep{williams2017mppi} optimize action sequences over fixed dynamics without deployment-time parameter updates, corresponding to single-loop learning \citep{Argyris1977DoubleLoop}: correcting actions without revising failure-producing assumptions. Model-free RL methods such as PPO \citep{schulman2017ppo} and SAC \citep{haarnoja2018sac} learn from scalar rewards across many episodes and resets, requiring reward engineering rather than single-episode adaptation. Model-based RL methods such as DreamerV3 \citep{hafner2023dreamerv3} learn world models but typically keep them fixed during deployment. In contrast, we implement double-loop learning, updating the agent's internal model within a single episode through language-diagnosed failure signals without resets or reward engineering.

\vspace{-1em}
\section{Reflective Test-Time Planning}
\vspace{-0.5em}

Consider an embodied agent with a multimodal large language model operating on task $\tau$ in a partially observable environment. At each timestep $t$, the model receives observation $o_t$, generates action $a_t$ in natural language, and receives execution feedback $e_t$ indicating whether the action executed successfully (e.g., object grasped, placement completed). Crucially, a positive $e_t$ indicates successful execution but does not imply the action was strategically correct or contributes to full task completion.

\begin{figure*}[t]\centering\includegraphics[width=\textwidth]{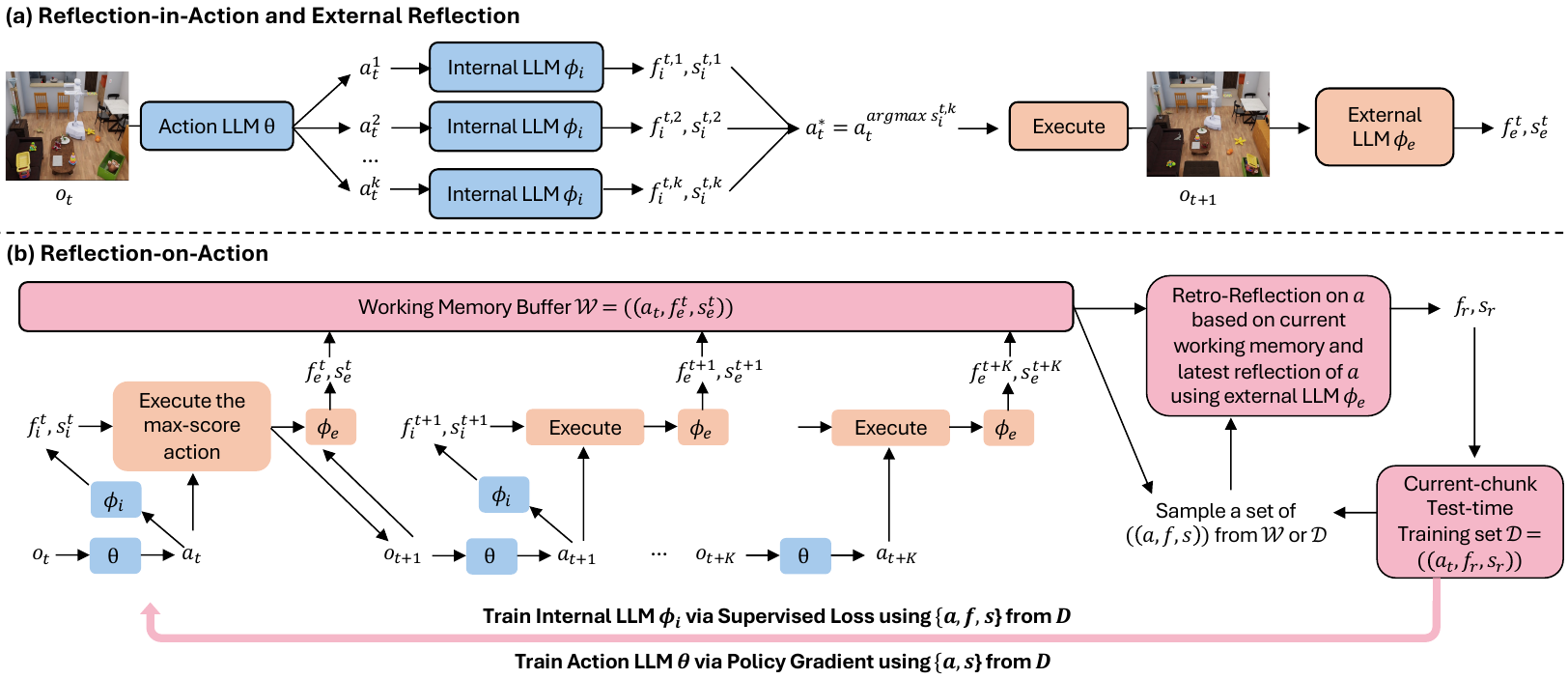}
\vspace{-1em}
\caption{\textbf{Method overview.} (a) \textit{Reflection-in-action}: multiple candidate actions are generated and scored by an internal reflection LLM prior to execution. (b) \textit{Reflection-on-action}: iteratively invoked when working memory hits K or at key milestones. Executed actions are critiqued by an external reflection LLM and stored in a working memory buffer; at milestones, hindsight re-evaluation assigns long-horizon credit. The resulting verbal reflections form self-supervised training data to update both the internal reflection LLM (supervised loss) and the action LLM (policy gradient) via test-time training, enabling agents to learn from execution experience during deployment.}
    \label{fig:method}
    \vspace{-0.5em}
\end{figure*}

% \vspace{-1em}
Traditional multimodal embodied LLM systems keep model parameters fixed at inference, limiting adaptation to novel scenarios or recovery from failures. We depart from static inference by building an adaptive test-time framework that employs three interacting models: an action generation LLM $\pi_\theta$ that produces actions given observations, an internal reflection LLM $V_{\phi_i}$ that generates pre-action evaluations, and an external reflection LLM $V_{\phi_e}$ that generates post-execution assessments. These LLMs are first initialized with basic capabilities for reasoning in  embodied environments through minimum supervised fine-tuning on a small set of tasks, enabling them to understand action formats, generate reflections, and process 3D observations before they can effectively learn from test-time experience. At deployment time, 
we introduce three reflection types: internal reflection $f_i$ for pre-action scoring, external reflection $f_e$ for post-execution assessment, and retrospective reflection $f_r$ for hindsight re-evaluation. We combine test-time scaling  (generating and scoring multiple candidate actions) for reflection-in-action, with test-time training (tuning $\pi_{\theta}$ and $V_{\phi_i}$) for reflection-on-action. Figure~\ref{fig:method} shows a method overview, and Algorithm~\ref{alg:reflective_ttt} provides a detailed method breakdown.

We choose verbal reflection as the representation for all reflection types inspired by ``double-loop learning'' \citep{Argyris1977DoubleLoop}: by articulating what went wrong and why, the agent abstracts generalizable lessons transferable to future decisions rather than reporting that an action failed. These articulated lessons serve as supervisory signals during deployment, providing interpretable feedback to be reused later. Thus, instead of only training the action model based on outcomes during test time (single loop),  we also train the internal reflection LLM to align its pre-action internal reflections with post-execution external reflections, updating the underlying reasoning process behind the action itself.

\vspace{-0.5em}
\subsection{Reflection-in-Action}
\vspace{-0.5em}
Humans naturally deliberate under uncertainty by \textit{mentally} simulating and reflecting on actions. We transfer this ability to embodied agents via \textit{reflection-in-action}: rather than greedily selecting the first plausible action, the agent samples candidates and reflects on each one before committing. We implement this through test-time scaling, where we generate $N$ diverse candidate actions and use the internal reflection LLM to produce reflective evaluations for each, which guide action selection.

\begin{algorithm}[t]
\fontsize{7pt}{7.2pt}\selectfont
\caption{Reflective Test-Time Planning}
\label{alg:reflective_ttt}
\begin{algorithmic}[1]
\REQUIRE Task $\tau$, initial observation $o_1$; action LLM $\pi_\theta$, internal reflection LLM $V_{\phi_i}$, external reflection LLM $V_{\phi_e}$; window size $K$
\STATE Initialize $a_0,f_e^{0}\leftarrow\text{None}$; buffers $\mathcal{W},\mathcal{D}_{\text{train}},\mathcal{D}_{\text{retro}}\leftarrow\emptyset$; temperature $T$; max step $\mathcal{T}$
\FOR{$t=1,\ldots,\mathcal{T}$}
    \STATE \textcolor{uclablue}{// Reflection-in-Action}
    \STATE Construct $x_{\text{action}}$ from $(\tau,o_t,a_{t-1},f_e^{t-1})$ and sample $\{a_t^k\}_{k=1}^N\sim\pi_\theta(\cdot|x_{\text{action}};T)$
    \FOR{$k=1,\ldots,N$}
        \STATE Construct $x_{\text{internal}}^k$ and score $f_i^{t,k},s_i^{t,k}\leftarrow V_{\phi_i}(x_{\text{internal}}^k)$
    \ENDFOR
    \STATE Select $a_t^*=a_t^{(\arg\max_k s_i^{t,k})}$; execute $a_t^*$ and observe $(o_{t+1},e_t)$; break if task complete

    \STATE \textcolor{uclablue}{// Reflection-on-Action}
    \STATE Construct $x_{\text{external}}$ and generate feedback $f_e^t,s_e^t\leftarrow V_{\phi_e}(x_{\text{external}})$
    \STATE $\mathcal{W}\leftarrow\mathcal{W}\cup\{(o_t,a_t^*,f_e^t,s_e^t)\}$

    \IF{$|\mathcal{W}|=K$ \OR hit key milestone}
        \STATE \textcolor{uclablue}{// Retrospective Reflection: Re-evaluate with Hindsight}
        \FOR{each $(o_j,a_j,f^j,s^j)\in\mathcal{W}\cup\mathcal{D}_{\text{retro}}$}
            \STATE Construct $x_{\text{retro}}^j$ from $(\tau,a_j,\mathcal{W},o_{t+1},f^j,s^j)$
            \STATE Generate $f_r^j,s_r^j\leftarrow V_{\phi_e}(x_{\text{retro}}^j)$ and set $\mathcal{D}_{\text{retro}}[a_j]\gets(a_j,f_r^j,s_r^j)$
        \ENDFOR
        \STATE $\mathcal{D}_{\text{train}}\leftarrow\mathcal{D}_{\text{retro}}$

        \STATE \textcolor{uclablue}{// Regularization: Prevent Catastrophic Forgetting}
        \FOR{sampled unexplored action $a_l$}
            \STATE Construct $x_{\text{internal}}^l$; get original output $f_i^l,s_i^l\leftarrow V_{\phi_i}(x_{\text{internal}}^l)$; add $(a_l,f_i^l,s_i^l)$ to $\mathcal{D}_{\text{train}}$
        \ENDFOR

        \STATE \textcolor{uclablue}{// Test-Time Training}
        \FOR{$(a,f,s_r)\in\mathcal{D}_{\text{train}}$}
            \STATE Construct $x_{\text{internal}},x_{\text{action}}$ from $a$; reward $r=2s_r/100-1$
            \STATE Update internal LLM: $\ell_\phi=-\log p_{\phi_i}(f|x_{\text{internal}})$, $\phi_i\leftarrow\phi_i-\eta_\phi\nabla_{\phi_i}\ell_\phi$
            \STATE Update action LLM: $\ell_\theta=-r\log p_\theta(a|x_{\text{action}})$, $\theta\leftarrow\theta-\eta_\theta\nabla_\theta\ell_\theta$
        \ENDFOR

        \STATE Clear working memory: $\mathcal{W}\leftarrow\emptyset$
    \ENDIF
\ENDFOR
\end{algorithmic}
\end{algorithm}

\textbf{Vanilla Action Generation.}  Standard autoregressive generation from the action LLM $\pi_\theta$ produces a single action via greedy or low-temperature sampling: 
\begin{equation}
a_t = \arg\max_{a \in \mathcal{A}} p_\theta(a | o_t)
\end{equation}
where $o_t$ is the current observation, with some architectures implicitly encoding historical context through memory structures (\textit{e.g.}, \cite{hu20253dllmmemlongtermspatialtemporalmemory}). Greedy generation commits to actions early without reflecting on potential consequences.

% \textbf{Candidate Generation.} At each decision step $t$, we sample $N$ diverse candidate actions using high-temperature sampling: 
% \vspace{-0.5em}
% \begin{equation}
% a_t^{k} \sim p_\theta(\cdot | o_t, a_{t-1}, f_e^{t-1}; T)\ \ \ k = 1, ..., N
% \vspace{-0.5em}
% \end{equation}

% where the model conditions on the current observation $o_t$, previous action $a_{t-1}$, and previous external reflection $f_e^{t-1}$ (both initialized as None; external reflection will be introduced later). High temperature $T$ encourages diversity. 

\textbf{Candidates Generation.} Different from the above, at each decision step $t$, we construct an action generation prompt $x_{\text{action}}$ containing task description $\tau$, current observation $o_t$, previous action $a_{t-1}$, and previous external reflection $f_e^{t-1}$ (both initialized as None; external reflection will be introduced later). We sample $N$ diverse candidate actions:
% \vspace{-0.5em}
\begin{equation}
a_t^{k} \sim p_\theta(\cdot | x_{\text{action}}; T)\ \ \ k = 1, ..., N
% \vspace{-0.5em}
\end{equation}
where high temperature $T$ encourages diversity in the generated candidates.

\textbf{Internal Reflection Scoring.} For each candidate, we construct an internal reflection prompt $x_{\text{internal}}^{k}$, which is identical to $x_{\text{action}}$ except that it adds the candidate action $a_t^{k}$ to be evaluated. The internal reflection LLM generates: \begin{equation}
f_i^{t,k}, s_i^{t,k} = V_{\phi_i}(x_{\text{internal}}^{k})
\end{equation}
where $s_i \in [0, 100]$ is a numerical score and $f_i^{t,k}$ is natural language reflection. Because this reflection occurs prior to action execution, we refer to it as internal reflection.

% \textbf{Internal Reflection Scoring.} For each candidate, we construct an internal reflection prompt $x_{\text{internal}}^{k}$ containing task description $\tau$, current observation $o_t$, previous action $a_{t-1}$, previous external reflection $f_e^{t-1}$, and candidate action $a_t^{k}$. The internal reflection LLM generates: \begin{equation}
% f_i^{t,k}, s_i^{t,k} = V_{\phi_i}(x_{\text{internal}}^{k})
% \end{equation}
% where $s_i \in [0, 100]$ is a numerical score and $f_i^{t,k}$ is natural language reflection. Because this reflection occurs prior to action execution, we refer to it as internal.

\textbf{Best Action Selection. }  We select the highest-scoring candidate:
\begin{equation}
a_t^* = a_t^{(\arg\max_{k \in [N]} s_i^{t,k})}
\end{equation}
 Rather than executing the first feasible action, the agent “mentally tries out” multiple options and chooses the one it internally judges as most promising.

\subsection{Reflection-on-Action}
\vspace{-0.5em}
Reflection-in-action has a  limitation: internal reflection operates in imagination, not reality. It may score an action highly based on plausible reasoning, yet the action fails due to unforeseen physical constraints or environmental dynamics. Reflection-on-action, learning from experience after actions are executed, addresses this by grounding learning in actual execution outcomes. 

\subsubsection{Multi-Scale External Reflection}
\textbf{External Reflection Generation.} After executing $a_t^{*}$ and observing $(o_{t+1}, e_t)$, we construct $x_{\text{external}}$ by extending $x_{\text{action}}$ with $a_t^{*}$, $e_t$, and $(o_t, o_{t+1})$.  The external reflection LLM generates:
\begin{equation}
f_e^{t}, s_e^{t} = V_{\phi_e}(x_{\text{external}})
\end{equation}
where $s_e^{t}$ is a score and $f_e^{t}$ is language feedback assessing the immediate outcome and its cause. This provides real-time assessment based on directly observable consequences.

\textbf{Working Memory Buffer.} We maintain a  buffer $\mathcal{W}_t = {(o_j, a_j, e_j, f_e^{j}) \mid j = t-K+1, ..., t}$ with $K$ steps. This buffer accumulates recent experience until reaching a key milestone (e.g., exiting a room, detecting repeated failures that need replanning) or when $|\mathcal{W}_t| = K$, at which point we trigger memory consolidation and test-time training.

\textbf{Retrospective Reflection with Hindsight.} A critical limitation of external reflection is that it evaluates actions based on immediate outcomes. An action may appear successful initially but later prove problematic (e.g., placing an object in an accessible compartment that blocks the only space for a larger object). To address this credit assignment problem, we introduce retrospective reflection.

Once we hit key milestone or reach the working memory limit, for each action $a_j$ that has been reflected before (either by immediate external reflection or by previous retro-reflection), the external reflection LLM re-evaluates the action with full hindsight:
\begin{equation}
f_r^{j}, s_r^{j} = V_{\phi_e}(x_{\text{retro}}^{j})
\end{equation}
where the retrospective prompt $x_{\text{retro}}^{j}$ includes: (1) the complete working memory window $\mathcal{W}_t$ as context; (2) a historical action $a_j$ to be retro-reflected based on the current outcomes; (3) its most recent reflection $f^{j}_{\text{recent}}$ (either $f_e^{j}$ from the current working memory $\mathcal{W}_t$ if this is the first retrospective evaluation, or $f_r^{j}$ from the previous retrospective round and stored in a retro-buffer $\mathcal{D}_{\text{retro}}$); and (4) the current observation $o_{t+1}$. After retro-revision, we update 
$\mathcal{D}_{\text{retro}}$ and store only the most recent retro-reflection for each action. As $\mathcal{W} \cup \mathcal{D}_{\text{retro}}$ grows larger with actions, we may subsample historical actions if necessary to keep it tractable.
% The retrospective score $s_r^{j} \in [0, 100]$ and reflection $f_r^{j}$ provide revised assessments incorporating knowledge of downstream consequences, and based on new information revealed by subsequent actions in the current working memory window.

\subsubsection{Test-Time Training Dataset Construction.} 
\vspace{-0.5em}
We construct training dataset $\mathcal{D}_{\text{train}}$ with two types of data:

\textbf{Retro-supervised pairs}: For any action $a_j$ that has been retrospectively evaluated, we create training pairs using the retrospective reflection $f_r^{j}$ and $s_r^{j}$: 
% where the input is an internal reflection prompt $x_{\text{internal}}^{j}$ and the target is the retrospective reflection $f_r^{j}$ and $s_r^{j}$ we obtained from the last part:
% \begin{equation}
% \mathcal{D}_{\text{retro}} = {(x_{\text{internal}}^{j} \rightarrow f_r^{j}, s_r^{j})}
% \end{equation}
\begin{equation}
\mathcal{D}_{\text{retro}} = {(a_j, f_r^{j}, s_r^{j})}
\end{equation}
% This trains the internal LLM $V_{\phi_i}$ to predict what future hindsight will reveal, learning to anticipate long-term consequences and preventing the model from generating incorrect internal reflections that lead to poor action selection.
These pairs use hindsight-corrected reflections and scores for training both the internal LLM $V_{\phi_i}$ and action LLM $\pi_{\theta}$.

\textbf{Regularization pairs}: 
To prevent catastrophic forgetting, we randomly sample unexplored actions $a_l$, construct $x_{\text{internal}}^{l}$ and use the internal LLM's current predictions:
\begin{equation}
\mathcal{D}_{\text{reg}} = {(a_l, f_i^{l}, s_i^{l})}
\end{equation}
% where $f_i^{l}, s_i^{l} = V{\phi_i}(x_{\text{internal}}^{l})$ are the model's current outputs.
% To prevent catastrophic forgetting, we randomly sample unexplored actions and use the internal feedback LLM's current predictions as targets:
% \begin{equation}
% \mathcal{D}_{\text{reg}} = {(x_{\text{internal}}^{l} \rightarrow f_i^{l}, s_i^{l})}
% \end{equation}
where $f_i^{l}, s_i^{l} = V_{\phi_i}(x_{\text{internal}}^{l})$ represents the model's current output for randomly sampled action $a_l$. This anchors the model to its existing knowledge for actions not updated by recent experience, preventing distribution shift caused by training exclusively on retrospectively evaluated actions.

\subsubsection{Test-Time Training}

\textbf{Internal Reflection LLM Training via Supervised Learning.}
We train the internal reflection LLM $V_{\phi_i}$ to predict retrospective reflections using standard supervised learning. The objective minimizes negative log-likelihood over the combined dataset $\mathcal{D}_{\text{train}} = \mathcal{D}_{\text{retro}} \cup \mathcal{D}_{\text{reg}}$:
\begin{equation}
\mathcal{L}_{\text{internal}}(\phi_{i}) = \mathbb{E}{(x_{\text{internal}}, f, s) \sim \mathcal{D}_{\text{train}}} [-\log p_{\phi_i}(f | x)]
\end{equation}
Where we construct $x_{\text{internal}}$ from each action $a$. We perform $E$ epochs of test-time training:
\begin{equation}
\phi_{i}^{(e+1)} = \phi_{i}^{(e)} - \eta_{\phi} \nabla_{\phi_{i}} \mathcal{L}_{\text{internal}}
\end{equation}

\textbf{Action LLM Training via RL.} The action LLM $\pi_\theta$ is updated using policy gradient with retrospective scores as rewards. We convert the retrospective score $s_r \in [0, 100]$ to a reward signal $
r = 2*(s_r / 100) - 1$
mapping scores to $[-1, 1]$.  
For each training example, we compute the log-probability of the executed action sequence:
\begin{equation}
\log p_\theta(a | x_\text{action}) = \sum_{i=1}^{|a|} \log p_\theta(a_i | a_{<i}, x_\text{action})
\end{equation}
where the sum is over action tokens.  We construct $x_{\text{action}}$ from each action $a$. 
The REINFORCE loss is:
\begin{equation}
\ell_{\theta} = -r \cdot \log p_\theta(a | x_\text{action})
\end{equation}
This gradient increases the probability of actions with positive rewards and decreases the probability of actions with negative rewards. We accumulate gradients over all examples in $\mathcal{D}_{\text{train}}$ over several RL\_steps:
\begin{equation}
\theta^{(s+1)} = \theta^{(s)} - \eta_{\theta} \nabla_{\theta} \sum_{(x_\text{action},f,s_r) \in \mathcal{D}_{\text{train}}} \ell_{\theta}(x_\text{action}, f, s_r)
\end{equation}

\section{Experiments on Long-Horizon Household Tasks}
\vspace{-0.5em}

\subsection{Long-Horizon Household Task Construction}
\vspace{-0.5em}
To evaluate our framework on tasks requiring multi-step reasoning and failure recovery, we construct Long-Horizon Household Tasks based on the BEHAVIOR-1K \citep{li2024behavior1k} environments. Inspired from household scenarios, we define four task categories: (1) \textbf{Fitting}, where objects must be packed or placed into constrained containers or surfaces, stressing geometry,  capacity and occlusion failures; (2) \textbf{Selection}, where the agent compares and retrieves the most suitable item (\textit{e.g.}, in terms of preferences or sizes). Failures occur when choices prove
suboptimal upon discovering better alternatives; (3) \textbf{Preparation}, where tasks require sequential constraints and dependencies (e.g., assembling or nested placement), stressing sequential dependency and non-local failures; and (4) \textbf{Hybrid}, where multiple modes appear within a single episode, stressing mixed spatial, relational, and occlusion failures.

\textbf{Task Generation \& Validation.}
We employ GPT-5 to generate task specifications through a structured 
prompting procedure using BEHAVIOR-1K scene graphs. Each generated task 
instance includes a task description, 3--7 relevant rooms, new objects 
with placement specifications, and a complete trajectory with interleaved 
actions, reflections, and scores. Since scene graphs provide object 
properties such as 3D bounding boxes, GPT-5 can deduce potential failures 
(e.g., size mismatches, occlusion) during generation. Generated trajectories 
are then executed in OmniGibson physics simulation to validate consistency 
with actual scene dynamics, rejecting instances where annotated outcomes 
conflict with simulator results. Full prompting and validation details are 
in Appendix~\ref{app:behavior}.

\begin{table*}[t]
\centering
\scriptsize
\setlength{\tabcolsep}{1.65pt}
\renewcommand{\arraystretch}{0.85}
\resizebox{\textwidth}{!}{%
\begin{tabular}{l|cccccc|ccccc|c}
\toprule
& \multicolumn{6}{c|}{\textbf{Baselines}} 
& \multicolumn{6}{c}{\textbf{Reflective Test-Time Planning Ablations}}\\
\midrule
&
Reflexion &
Self-Refine &
ReflectVLM &
PPO &
DreamerV3 &
3DLLM-Mem &
w/o RIA/ROA &
w/o RIA &
w/o ROA &
w/o Act. &
w/o Int. &
\textbf{Ours} \\
\midrule
Fitting     & 8.51\% & 10.6\% & 2.12\% & 0\% & 4.26\% & 10.6\% & 0\%    & 33.5\% & 6.38\% & 25.5\% & 12.8\% & \textbf{44.7\%} \\
Selection   & 8.82\% & 11.8\% & 5.88\% & 2.94\% & 11.8\% & 14.7\% & 17.6\% & 5.88\% & 11.8\% & 26.5\% & 8.82\% & \textbf{32.4\%} \\
Preparation & 15.9\% & 12.7\% & 14.3\% & 7.94\% & 11.1\% & 9.52\% & 11.1\% & 3.17\% & 19.0\% & 20.6\% & 17.5\% & \textbf{31.7\%} \\
Hybrid      & 6.45\% & 9.68\% & 6.45\% & 3.23\% & 12.9\% & 9.68\% & 12.9\% & 3.23\% & 12.9\% & 16.1\% & 9.68\% & \textbf{25.8\%} \\
\midrule
Average & 9.92\% & 11.20\% & 7.19\% & 3.53\% & 10.02\% & 11.13\% & 10.40\% & 11.45\% & 12.52\% & 22.18\% & 12.20\% & \textbf{33.65\%} \\
\bottomrule
\end{tabular}%
}
\vspace{-0.5em}
\caption{Baseline comparisons and ablations of Reflective Test-Time Planning on long-horizon household tasks. RIA denotes reflection-in-action and ROA denotes reflection-on-action. Act. denotes action model loss; Int. denotes internal reflection model loss.}
\vspace{-2.0em}
\label{tab:behavior}
\end{table*}
\textbf{Finetuning, Evaluation \& Implementation.}
Each validated trajectory yields (observation, action, reflection, score) 
tuples that form SFT pairs to initialize our three embodied LLMs before 
test-time deployment. Given SFT data, we build on LLaVA-3D~\citep{zhu2024llava3d} and 
train a single unified model on all three modes (action generation, 
internal reflection, external reflection) for cross-mode learning, then 
instantiate three copies at deployment as $\pi_\theta$, $V_{\phi_i}$, 
and $V_{\phi_e}$. During evaluation, the agent is given only the scene 
configuration and task description and must autonomously generate and 
execute its trajectory step by step. Finetuning and evaluation sets have 
no overlap in task descriptions, scene configurations, or object 
placements. A task succeeds if all required objects reach target locations 
within the action budget. Full model architecture, training hyperparameters, 
and baseline implementation details are provided in 
Appendix~\ref{behavior-details}-~\ref{baselines}.

\vspace{-0.5em}
\begin{table*}[h]
\centering
\begin{minipage}{0.46\textwidth}
\centering
\setlength{\tabcolsep}{2pt}
\renewcommand{\arraystretch}{0.7}
\scriptsize
\begin{tabular}{lcccc}
\toprule
\textbf{\#Data} & \textbf{1 Epoch} & \textbf{2 Epoch} & \textbf{3 Epoch} & \textbf{w/o RIA/ROA} \\
\midrule
25\%          & 15.6 & 20.3 & 22.1 & 1.3 \\
50\%          & 18.7 & 24.8 & 19.6 & 4.4 \\
100\% (Ours)  & 33.6 & 20.7 & 16.1 & 10.4 \\
200\%         & 24.5 & 19.6 & 9.1  & 13.4 \\
300\%         & 20.7 & 17.5 & 7.5  & 15.9 \\
\bottomrule
\end{tabular}
\vspace{-0.5em}
\caption{SFT scaling analysis. Too little data fails to produce correct formats; too much overfits to SFT distribution, destabilizing TTT.}
\label{tab:sft_scaling_main}
\end{minipage}
\hfill
\begin{minipage}{0.52\textwidth}
\centering
\setlength{\tabcolsep}{3pt}
\scriptsize
\begin{tabular}{lcccccc}
\toprule
\textbf{Method} & \textbf{Fit} & \textbf{Sel.} & \textbf{Prep.} & \textbf{Hyb.} & \textbf{Avg.} & \textbf{Overhead} \\
\midrule
TTT on Ext.\ Refl.   & 22.4 & 17.6 & 12.7 & 9.69 & 15.6 & $\sim$4$\times$ \\
TTT on Retro (Ours)  & 44.7 & 32.4 & 31.7 & 25.8 & 33.7 & 1$\times$ \\
\bottomrule
\end{tabular}

\caption{Retrospective vs.\ immediate external reflection as TTT signal. Retro-reflection achieves better credit assignment at lower overhead, as external reflection triggers at every step.}
\label{tab:retro_vs_ext_main}
\end{minipage}
\end{table*}
\vspace{-1.5em}
\subsection{Experimental Results \& Analysis}
\vspace{-0.5em}
\paragraph{Results Analysis.}
Table~\ref{tab:behavior} shows substantial improvements of our model over both 
baselines and ablations across all task categories. Fitting tasks 
benefit most (44.7\% vs.\ 10.6\% for 3DLLM-Mem, 2.1\% for ReflectVLM, 
and 0\% for PPO), as their tight spatial constraints demand iterative 
refinement and continuous adjustment of 3D geometric understanding 
based on execution feedback. The ablation studies reveal that RIA and ROA are mutually dependent: 
removing either causes performance degradation, and sometimes removing 
just one component performs worse than removing both. Without RIA, 
Preparation falls to 3.17\% and Hybrid to 3.23\%, below the 11.1\% 
and 12.9\% achieved by removing both components. RIA without ROA 
produces overconfident yet inaccurate action scores with no hindsight 
correction, while ROA without RIA wastes learning on poorly chosen 
actions that fail to reveal true scene affordances. Together they form 
a virtuous cycle: best-of-N selection yields higher-quality trajectories 
for learning, and TTT refines the internal model, leading to better 
future action selection. The loss ablations confirm the same mutual 
dependency between the action loss and internal reflection loss: without 
the internal loss, Hybrid drops to 9.68\% and Selection to 8.82\%, both 
below their respective w/o ROA baselines. Table~\ref{tab:hm3d} shows zero-shot generalization to photorealistic 
HM3D environments. Our model outperforms all baselines by a large 
margin despite training exclusively on synthetic BEHAVIOR-1K scenes. 
The qualitative example in Figure~\ref{fig:qualitative}(a) further 
illustrates continual learning and active perception driven by 
accumulated scene experience. Test-time cost analysis in 
Appendix~\ref{appendix:test_time_compute} shows even with 
substantially more test-time compute, baselines continue to repeat 
failures. Retro-reflection quality, search-based baselines, efficiency improvements,
commercial LLM baselines, and generalization to human instructions are in 
Appendices~\ref{appendix:efficiency}, \ref{appendix:retro_quality}--~\ref{appendix:commercial_llms}.

\vspace{-1em}
\paragraph{Necessity and Reliability of SFT Initialization and External LLM.}
A key design question is how much our framework relies on the quality 
of GPT-5-generated initialization data and the fixed external LLM. 
Table~\ref{tab:sft_scaling_main} analyzes the role of SFT initialization 
and establishes three points.
(1) SFT is necessary but minimal.
GPT-5-generated data serves only as a minimal bootstrap: it teaches 
the model to produce structured action formats and reflections, not to 
solve tasks. Without any initialization, test-time reflection yields 
near-zero performance regardless of whether few-shot in-context examples 
are provided, confirming necessity (Appendix~\ref{appendix:sft_analysis}); 
the inverted-U pattern shows that too little data produces malformed 
outputs while too much overfits to GPT-5's reasoning distribution and 
destabilizes TTT.
(2) Gains come from reflection, not initialization.
The initialization-only variant achieves only 10.4\% average success 
rate against 33.7\% for the full model. While GPT-5 provides format, 
the reasoning, recovery, and adaptation abilities that drive performance 
are acquired entirely through test-time interaction with the environment.
(3) The external LLM provides a reliable and independent learning signal.
Table~\ref{tab:sft_scaling_main} shows that even with minimal SFT 
initialization, $V_{\phi_e}$ consistently drives meaningful performance 
gains across all data fractions and epoch counts. This is because 
$V_{\phi_e}$ acts as a post-hoc describer of directly observed outcomes 
grounded in $e_t$ from the simulator, unlike $\pi_\theta$ and 
$V_{\phi_i}$ which must reason about unobserved futures. Human annotators 
confirm high factual correctness, causal quality, and usefulness of GPT-5  data and external reflection LLM for even 
out-of-domain scenarios (Appendix~\ref{appendix:vphie_reliability}\&\ref{appendix:gpt5_quality}). However, 
Table~\ref{tab:retro_vs_ext_main} shows that applying this signal at 
every step yields only 15.6\% at $\sim$4$\times$ overhead, demonstrating 
that reliable description alone is insufficient without long-horizon 
credit assignment via retrospective reflection at key milestones.

\begin{figure*}[t]\centering\includegraphics[width=0.95\textwidth]{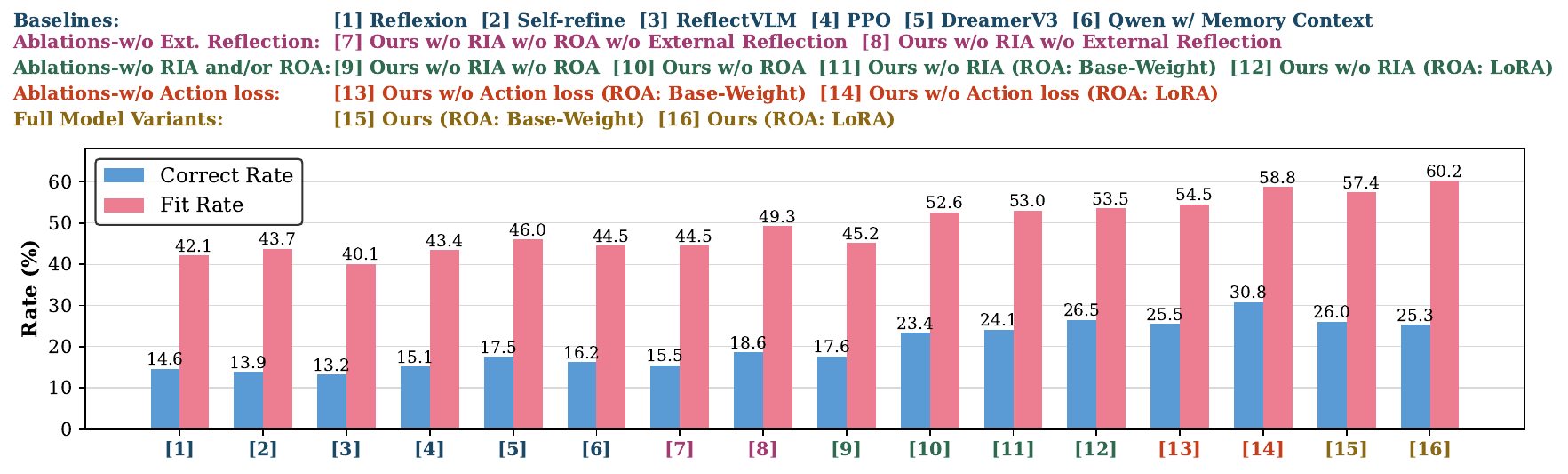}
\vspace{-1em}
\caption{\textbf{Cupboard Fitting results}. Blue bars show correct-placement rate, pink bars show fit rate. RIA means Reflection-in-action; ROA means Reflection-on-action. ``W/o external reflection": we don't use external reflection as the input to the action generation LLM. We implement two test-time training variants for ROA: 1) test-time training on all base weights; 2) test-time training on LoRA parameters only. Reflective Test-Time Planning significantly improves both success metrics.}
\vspace{-1.5em}
\label{fig:cupboard_result}
    \label{fig:cupboard}
\end{figure*}

\vspace{-1em}
\section{Experiments on the Cupboard Fitting Task}
\vspace{-0.5em}
\begin{wrapfigure}{l}{0.4\linewidth}
    \vspace{-1.0em}
    \centering
    \includegraphics[width=\linewidth]{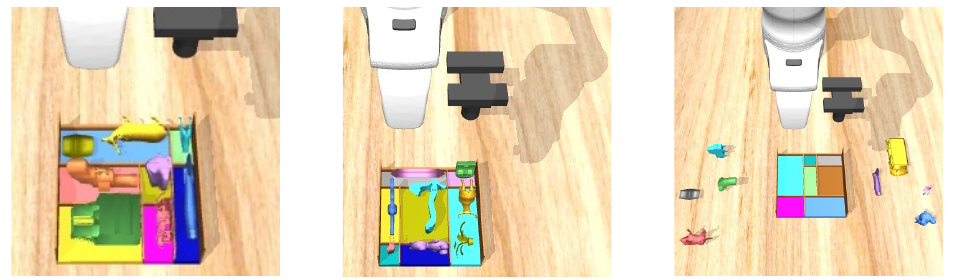}
    \vspace{-2em}
    \caption{The Cupboard Fitting Task.}
    \label{fig:cupboard}
    \vspace{-1em}
\end{wrapfigure}

\vspace{-0.5em}
\textbf{Cupboard Fitting Task Design.}
The Cupboard Fitting task serves as a complementary benchmark to 
Long-Horizon Household Tasks. While BEHAVIOR provides realistic 
multi-room environments with complex semantic reasoning challenges, 
it also introduces environmental uncertainties that make it difficult 
to isolate learning mechanisms. We design Cupboard Fitting as a 
controlled MuJoCo environment where agents learn from placement 
failures, enabling precise measurement of reflective test-time 
training mechanisms. The environment features a multi-compartment 
cupboard and a set of colored geometric objects that a Franka Panda 
arm must place via high-level natural language commands. Success 
requires reasoning about object-compartment compatibility, multi-object 
spatial packing, and long-horizon dependencies where early placement 
decisions affect later possibilities. We define two metrics: 
\textit{fit rate} measures objects successfully placed in any 
compartment, and \textit{correct rate} measures objects placed in 
their designated compartments. Full environment and implementation 
details are in Appendix~\ref{appendix:cupboard_details}.

\textbf{Experimental Results \& Analysis.}
Figure~\ref{fig:cupboard_result} shows that our full method achieves 
60.2\% fit rate and 25.3\% correct rate, substantially outperforming 
all baselines. Both reflection mechanisms are essential: removing RIA 
drops performance to 53.5\% and removing ROA to 45.2\% fit rate, with 
removing both degrading further to 44.5\%. LoRA-based TTT (60.2\%) 
performs comparably to full base-weight training (57.4\%) while reducing 
trainable parameters, confirming that reflection-in-action 
for candidate selection and reflection-on-action for test-time adaptation 
together enable effective learning from failures during deployment.
% In Figure~\ref{fig:qualitative}(b), we show a qualitative result where we generalize our model in the real-robot setting.
% % demonstrate that our approach generalizes beyond simulation to real-robot execution (Figure~\ref{fig:qualitative}(b)). 
% In these trials, reflective adaptation enables the robot to recover from execution failures, avoid repeated placement errors, and correct earlier decisions through retrospective reflection, illustrating that the learned behaviors transfer to the physical world with satisfying generalization abilities. 
Appendix~\ref{sec:receding} also shows that our method enables long-horizon planning by retrospective reflection that outperforms Receding Horizon Planning while saving 5$\times$ test-time compute. Hyperparameter analyses of both ROA \& RIA are provided in Appendix~\ref{sec:hyperparameter_analysis}.

\vspace{-1em}
\begin{figure*}[htbp]\centering\includegraphics[width=0.95\textwidth]{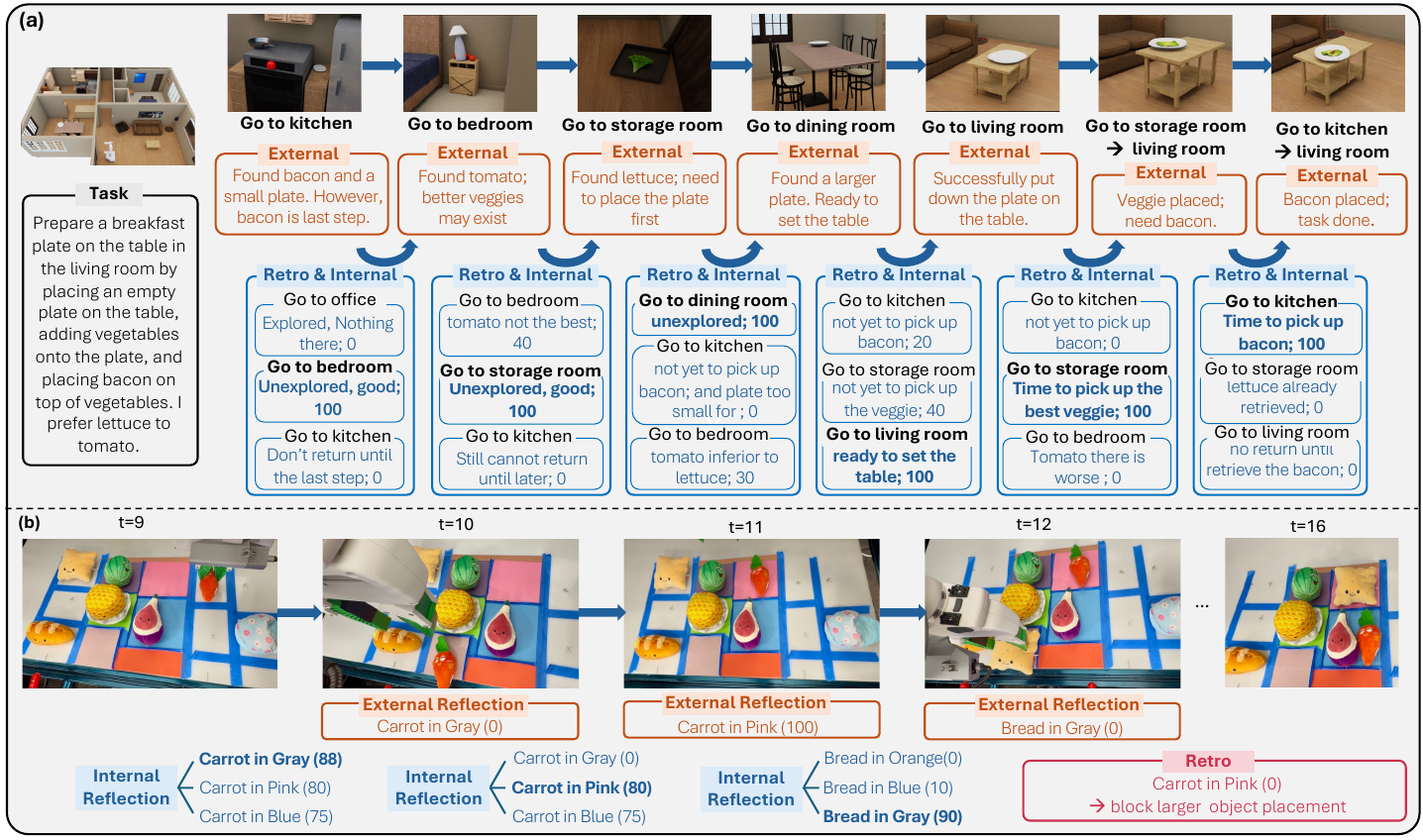}
\vspace{-0.5em}
\caption{\textbf{Qualitative Examples.} Steps and reflections simplified for better presentations. Blue text: internal reflection. Orange text: external reflection. red text: retrospective reflection.  (a) Long-Horizon Household example. We use retro \& internal because the generated retro reflection is also used to train the internal model.  (b) Real-robot Cupboard Fitting example. We put reflection scores inside brackets, omit detailed reflections and only present the scores for simplicity.}
\vspace{-0.5em}
    \label{fig:qualitative}
\end{figure*}

\begin{table*}[h]
\centering
\scriptsize
\setlength{\tabcolsep}{4pt}
\renewcommand{\arraystretch}{0.8}
\begin{minipage}{0.46\textwidth}
\centering
\begin{tabular}{lcc}
\toprule
\textbf{Setting} & \textbf{Fit \%} & \textbf{Correct \%} \\
\midrule
Zero-shot, w/o RIA \& ROA  & 13.1 & 4.5  \\
Fine-tuned, w/o RIA \& ROA & 20.7 & 7.2  \\
Zero-shot, full            & 40.6 & 13.4 \\
Fine-tuned, full           & 44.2 & 16.6 \\
\bottomrule
\end{tabular}
\vspace{-0.5em}
\caption{Real-robot Cupboard Fitting on a Franka Panda arm. Zero-shot denotes direct sim-to-real transfer.}
\label{tab:real_robot}
\end{minipage}
\hfill
\begin{minipage}{0.46\textwidth}
\centering
\begin{tabular}{lc}
\toprule
\textbf{Method} & \textbf{Preparation (\%)} \\
\midrule
Reflexion   & 2.44 \\
Self-Refine & 4.88 \\
ReflectVLM  & 0.00 \\
PPO         & 0.00 \\
DreamerV3   & 2.44 \\
3DLLM-Mem   & 7.32 \\
\textbf{Ours} & \textbf{19.5} \\
\bottomrule
\end{tabular}
\vspace{-0.5em}
\caption{Zero-shot generalization to HM3D.}
\label{tab:hm3d}
\end{minipage}
\end{table*}

\vspace{-1em}

\paragraph{Real-Robot Experiments.}
Table~\ref{tab:real_robot} reports results on a Franka Panda arm under 
both zero-shot and fine-tuned settings. The full 
model generalizes substantially better than the ablation without RIA and 
ROA, with the performance gap widening in the real-world setting: 
real-world uncertainties such as grasp imprecision and object slippage 
introduce the kind of unexpected failures that reflection-on-action 
is designed to recover from. Notably, even without any 
fine-tuning, the full model  outperforms the zero-shot fine-tuned 
ablation, suggesting that active failure recovery matters more 
than domain adaptation when execution errors are the primary bottleneck. 
The qualitative example in Figure~\ref{fig:qualitative}(b) further 
illustrates how retrospective reflection enables the robot to correct 
earlier placement decisions mid-episode. Full real-robot setup details are provided in Appendix~\ref{appendix:real_robot}.

\vspace{-0.5em}
\section{Conclusion}
\vspace{-0.5em}
We introduce Reflective Test-Time Planning, which couples reflection-in-action for pre-action evaluation with reflection-on-action for post-execution assessments.
Evaluations across two newly-designed embodied tasks demonstrate strong gains and highlight the complementary roles of the two reflection modes. Future work may extend reflective adaptation to richer sensory modalities (\textit{e.g.}, tactile).

\section*{Acknowledgments}
This paper is funded by ONR MURI N00014-24-1-2748,  also supported by IITP funded by the Korean Government (MSIT) (No. RS-2024-00457882, National AI Research Lab Project), and Grant AW1134392 (Reasoning in Motion) from the TRI University 3.0 Program. This work used computational resources provided by Google through the Google Gemini Academic Program.
% \section*{Impact Statement}
% The positive impact includes more robust household robots that recover from mistakes, enabling safer deployment in unstructured environments. However, autonomous behavior updates during deployment raise important considerations: agents might develop unexpected strategies that bypass safety constraints, verbal reflections could inherit language model biases, and improved failure recovery may reduce human oversight in safety-critical applications. We believe transparency through interpretable verbal reflections and careful monitoring during initial deployments can help mitigate these risks while advancing more capable and trustworthy embodied AI systems.

\bibliography{example_paper}
\bibliographystyle{plainnat}

\newpage
\appendix
\appendix
\clearpage

{
    \hypersetup{linkcolor=black}
    \tableofcontents
}
\clearpage

\section{Contribution Statement}
Yining Hong proposed the idea; implemented all codes, data, experiments and wrote the paper. Huang Huang helped set up the real-world robot. The other authors contributed through advisory support and research supervision.

\section{Broader Impacts}
\label{appendix:broader_impacts}

The positive impact includes more robust household robots that recover from mistakes, enabling safer deployment in unstructured environments. However, autonomous behavior updates during deployment raise important considerations: agents might develop unexpected strategies that bypass safety constraints, verbal reflections could inherit language model biases, and improved failure recovery may reduce human oversight in safety-critical applications. We believe transparency through interpretable verbal reflections and careful monitoring during initial deployments can help mitigate these risks while advancing more capable and trustworthy embodied AI systems.

\section{Test-Time Cost Analysis}
\label{appendix:test_time_compute}

\subsection{Computational Comparison}
Our full model introduces two additional sources of inference time cost:
(i) candidate action sampling and internal reflection scoring for reflection-in-action (RIA), and
(ii) periodic reflection-on-action(ROA)-based LoRA test-time updates.

On average across Long-Horizon Household Tasks and Cupboard Fitting Tasks, we observe a $\sim 3\times$ increase in per-step wall-clock time compared to the vanilla baseline:
\[
\text{Time}_{\text{full}} \approx 3.0 \times \text{Time}_{\text{no-RIA/ROA}}.
\]
% The majority of overhead arises from internal reflection scoring, while ROA contributes amortized cost during milestone-triggered updates. 
Importantly, this latency is incurred at deployment and does not require additional supervised data or environment rollouts beyond normal task execution.

\subsection{Why the Overhead is Justified}
Despite the additional test-time latency, the reflective overhead is justified for three reasons:

\paragraph{Deployment-Oriented Adaptation.}
The extra time cost occurs at deployment rather than pretraining. This aligns with realistic embodied settings where robots adapt online rather than relying on costly retraining cycles.

\paragraph{Reduction of Execution Waste.}
Vanilla agents frequently repeat failures (e.g., placing incompatible objects, revisiting rooms without intent). RIA reduces execution waste by filtering poor actions before execution, while ROA eliminates repeated failures through hindsight-driven updates. A smaller number of higher-quality actions amortizes the reflection cost.

\paragraph{Conversion of Time into Learning.}
Extra time is not spent on naive rollouts, but on improving internal models. RIA and ROA produce persistent behavioral improvements, whereas baselines spend time without updating policies or reasoning mechanisms. In short, reflective time is structurally more valuable than mere rollout time.

\begin{table}[htb]
\centering
\small
\setlength{\tabcolsep}{4pt}
\begin{tabular}{l|cc|c}
\hline
 & \multicolumn{2}{c|}{\textbf{w/o RIA w/o ROA}} & \textbf{Ours (Full)} \\
\hline
 & \textbf{Vanilla} & \textbf{Same Time Budget} & \\
\hline
Fitting     & 0.00\%   & 0.00\%   & 44.7\% \\
Selection   & 17.6\%   & 14.7\%   & 32.4\% \\
Preparation & 11.1\%   & 12.7\%   & 31.7\% \\
Hybrid      & 6.45\%   & 6.45\%   & 25.8\% \\
\hline
Average     & 8.79\%   & 8.46\%   & 33.65\% \\
\hline
\end{tabular}
\caption{Performance comparison between (1) vanilla ablation without RIA or ROA, (2) vanilla baseline with matched time budget (3× steps), and (3) our full reflective model on Long-Horizon Household Tasks.}
\label{tab:compute_matched_baseline}
\end{table}

\subsection{Compute-Matched Experiment}
A natural concern is whether the performance gap is merely due to increased wall-clock time. To evaluate this, we construct a time-matched variant where the vanilla baseline receives a $3\times$ steps budget, matching the approximate inference time of our full model:
\[
\text{Steps}_{\text{baseline}}^{\text{expanded}} \approx 3\times \text{Steps}_{\text{baseline}}.
\]

From Table \ref{tab:compute_matched_baseline}, we observe that even under a tripled time budget, the baseline:

\begin{itemize}
    \item fails to correct early suboptimal decisions,
    \item frequently revisits states without strategic change,
    \item exhibits repeated placement/navigation failures,
    \item plateaus significantly below our full model on all benchmarks and does not improve over the vanilla baseline. 
\end{itemize}

This result supports that reflective time is not equivalent to naive rollout time expansion: reflective updates change the decision process itself, while rollout expansion merely increases trajectory length without improving competence or hindsight reasoning.

\paragraph{Conclusion.}
These findings indicate that although our method incurs a $\sim 3\times$ test-time latency overhead, reflective computation provides unique adaptation benefits that cannot be recovered by proportional rollout-step scaling. The overhead is therefore practical and justified in embodied deployment settings.

\section{Efficiency Improvements via Parallelization}
\label{appendix:efficiency}

We explored several strategies to reduce the $\sim3\times$ overhead without sacrificing performance. Results are shown in Table~\ref{tab:efficiency}.

\begin{table}[htb]
\centering
\small
\setlength{\tabcolsep}{6pt}
\begin{tabular}{lcc}
\toprule
\textbf{Configuration} & \textbf{Success} & \textbf{Overhead} \\
\midrule
Ours (full)                      & 33.7\% & 1.00$\times$ \\
+Parallel Scoring + KV Cache     & 33.0\% & 0.85$\times$ \\
+Batched Retro-Reflection        & 33.5\% & 0.66$\times$ \\
+TTT Early-Stop                  & 32.3\% & 0.60$\times$ \\
+QLoRA                           & 32.0\% & 0.52$\times$ \\
+Multi-GPU Parallelism           & 31.5\% & 0.45$\times$ \\
\bottomrule
\end{tabular}
\caption{Cumulative efficiency improvements. Each row adds to the previous. The combined pipeline reduces overhead to $0.45\times$ with minimal performance degradation.}
\label{tab:efficiency}
\end{table}

\textbf{Parallel candidate scoring + KV cache.} Candidate scoring calls are independent given the shared prefix context and can run in parallel with a shared KV cache.
\textbf{Batched retro-reflection.} Retro-reflections in the working memory window can be batched in a single forward pass.
\textbf{TTT early-stop.} Training halts when the loss converges rather than running for a fixed number of epochs.
\textbf{QLoRA.} 4-bit quantized LoRA reduces TTT memory footprint with minimal performance degradation.
\textbf{Multi-GPU parallelism.} Model parallelism, distributed LoRA updates, and asynchronous TTT updates on a separate GPU further reduce wall-clock time.

Together these techniques reduce overhead from $\sim3\times$ to $\sim0.45\times$ while maintaining 31.5\% success rate, demonstrating that the framework is practically deployable with engineering effort.

\section{Generalization to Habitat-Matterport 3D Scenes: More Details}
\label{sec:habitat_generalization}

\subsection{Experimental Setup}

To evaluate the generalization capacity of our Reflective Test-Time Planning framework, we conduct additional experiments on the Habitat-Matterport 3D (HM3D) dataset~\citep{ramakrishnan2021habitatmatterport3ddatasethm3d}, which provides photorealistic 3D reconstructions of real-world indoor environments. Unlike BEHAVIOR-1K scenes which are primarily synthetic household environments, HM3D offers diverse real-world scenes with different spatial layouts, object distributions, and visual appearances, providing a challenging domain shift for testing our method's transferability.

We focus specifically on \textbf{Preparation tasks}, which involve sequential constraints and dependencies where actions must occur in specific orders. These tasks are particularly challenging in HM3D environments due to: (1) more complex spatial layouts with irregular room configurations, (2) diverse object appearances and placements not seen during training, (3) ambiguous spatial relationships that require active exploration to resolve, and (4) longer navigation distances between task-relevant objects.

\subsection{Task Construction and Adaptation}

We adapt our task generation pipeline to HM3D scenes while maintaining the same core task structure. We select 41 preparation task instances across diverse HM3D scenes, ensuring coverage of various spatial configurations and object arrangements. Each task requires the agent to: (1) navigate through multiple rooms to locate task-relevant objects, (2) retrieve objects in the correct sequential order, (3) perform placement or assembly actions with proper dependencies, and (4) handle spatial constraints specific to real-world scene layouts.

The key difference from Long-Horizon Household evaluation is the domain gap: our models are trained exclusively on BEHAVIOR-1K synthetic scenes and must generalize to HM3D's photorealistic environments at test time. This tests whether reflection-based adaptation can overcome distribution shift through deployment-time learning.

\subsection{Implementation Details}

We use the same model architecture (LLaVA-3D-7B) and test-time training configuration as the Long-Horizon Household experiments, with no additional fine-tuning on HM3D scenes. Point cloud observations are extracted from RGB-D sensors in the same manner, and the reflection generation prompts remain unchanged. This zero-shot transfer setup provides a rigorous test of whether reflective mechanisms learned in synthetic environments transfer to real-world scene understanding.

% \subsection{Results}

% \begin{table}[htb]
% \centering
% \small
% \setlength{\tabcolsep}{4pt}
% \begin{tabular}{lc}
% \toprule
% \textbf{Method} & \textbf{Preparation (\%)} \\
% \midrule
% Reflexion          & 2.44 \\
% Self-Refine        & 4.88 \\
% ReflectVLM         & 0.00 \\
% PPO                & 0.00 \\
% DreamerV3          & 2.44 \\
% 3DLLM-Mem          & 7.32 \\
% \midrule
% w/o RIA, w/o ROA   & 0.00 \\
% w/o ROA            & 7.32 \\
% w/o RIA            & 0.00 \\
% w/o Act.\ Loss     & 14.6 \\
% w/o Int.\ Loss     & 9.76 \\
% \midrule
% \textbf{Ours}      & \textbf{19.5} \\
% \bottomrule
% \end{tabular}
% \caption{Generalization to HM3D (41 Preparation tasks, zero-shot from BEHAVIOR-1K). Our framework maintains substantial relative advantage over all baselines despite domain shift.}
% \label{tab:habitat}
% \end{table}

% Our method achieves 19.5\% success rate despite zero-shot transfer from synthetic to photorealistic environments, maintaining over 60\% of its BEHAVIOR performance (31.7\% $\to$ 19.5\%) and substantially outperforming all baselines. Several methods collapse to 0\% (ReflectVLM, PPO, w/o RIA w/o ROA, w/o RIA), highlighting that the combination of RIA and ROA is critical for robustness under domain shift.

\section{Reliability of the External Reflection Model}
\label{appendix:vphie_reliability}

\subsection{Human Evaluation of $V_{\phi_e}$ Quality}

Human annotators rate $V_{\phi_e}$ outputs on three dimensions: Factual Correctness, Causal Quality, and Usefulness as a learning signal.

\begin{table}[htb]
\centering
\small
\setlength{\tabcolsep}{6pt}
\begin{tabular}{lccc}
\toprule
\textbf{Setting} & \textbf{Factual} & \textbf{Causal} & \textbf{Useful} \\
\midrule
In-domain (BEHAVIOR) & 98 & 95 & 99 \\
Out-of-domain (HM3D) & 96 & 96 & 95 \\
\bottomrule
\end{tabular}
\caption{Human evaluation of $V_{\phi_e}$ reflection quality (\%). Quality remains high under distribution shift.}
\label{tab:vphie_human}
\end{table}

\subsection{Human Oracle Comparison}

To directly isolate $V_{\phi_e}$'s reliability as a supervision source, we replace $V_{\phi_e}$'s reflections with human-written reflections of the same execution outcomes on the fly, serving as an oracle upper bound. We evaluate on HM3D (environment-level shift) and two held-out BEHAVIOR task types (task-level shift).

\begin{table}[htb]
\centering
\small
\setlength{\tabcolsep}{6pt}
\begin{tabular}{lccc}
\toprule
\textbf{Method} & \textbf{HM3D} & \textbf{Cleanup} & \textbf{Rearrange} \\
\midrule
Ours w/ $V_{\phi_e}$     & 19.5\% & 40.0\% & 26.7\% \\
Ours w/ Human Oracle      & 21.9\% & 40.0\% & 33.3\% \\
\bottomrule
\end{tabular}
\caption{$V_{\phi_e}$ vs.\ human oracle under distribution shift. Near-identical performance confirms $V_{\phi_e}$ provides supervision of equivalent quality to human judgment.}
\label{tab:oracle}
\end{table}

The near-identical performance confirms that $V_{\phi_e}$ is a reliable teacher signal under both environment-level and task-level shift. This is expected: $V_{\phi_e}$ serves as a post-hoc describer of directly observed outcomes grounded in a binary execution signal $e_t$ from the simulator, not a forward predictor reasoning about unobserved futures.

\section{SFT Initialization Analysis}
\label{appendix:sft_analysis}

% \subsection{SFT Scaling Analysis}

% \begin{table}[htb]
% \centering
% \small
% \setlength{\tabcolsep}{4pt}
% \begin{tabular}{lcccc}
% \toprule
% \textbf{\#Data} & \textbf{1 Epoch} & \textbf{2 Epochs} & \textbf{3 Epochs} & \textbf{w/o RIA/ROA} \\
% \midrule
% 25\%          & 15.6 & 20.3 & 22.1 & 1.3 \\
% 50\%          & 18.7 & 24.8 & 19.6 & 4.4 \\
% 100\% (Ours)  & 33.6 & 20.7 & 16.1 & 10.4 \\
% 200\%         & 24.5 & 19.6 & 9.1  & 13.4 \\
% 300\%         & 20.7 & 17.5 & 7.5  & 15.9 \\
% \bottomrule
% \end{tabular}
% \caption{SFT scaling analysis (success rate \%, w/ RIA/ROA unless noted). 100\% data, 1 epoch is the minimal bootstrap: less data fails to produce correct formats; more data or epochs overfits to GPT-5's distribution, making TTT unstable.}
% \label{tab:sft_scaling}
% \end{table}

\subsection{Zero-SFT Baselines}

\begin{table}[htb]
\centering
\small
\setlength{\tabcolsep}{4pt}
\begin{tabular}{lc}
\toprule
\textbf{Method} & \textbf{Accuracy (\%)} \\
\midrule
Zero-SFT, w/o RIA \& ROA                        & 0.0  \\
Zero-SFT, w/ RIA \& ROA                         & 0.0  \\
Zero-SFT, Few-Shot In-Context, w/o RIA \& ROA   & 1.3  \\
Zero-SFT, Few-Shot In-Context, w/ RIA \& ROA    & 2.1  \\
\midrule
Ours, w/o RIA \& ROA                            & 10.4 \\
Ours (Full)                                      & 33.7 \\
\bottomrule
\end{tabular}
\caption{Zero-SFT baselines confirm that test-time reflection cannot function without minimal SFT initialization.}
\label{tab:zero_sft}
\end{table}

\subsection{Base Model Scale Analysis}

Table~\ref{tab:model_scale} analyzes the minimal level of embodied 
prior required for reflection to produce semantically reliable signals, 
across three model scales. We measure four dimensions: semantic 
similarity (cosine similarity between model output and SFT ground-truth, 
assessing whether reflections are semantically meaningful), human 
evaluation (annotator ratings of reflection quality), format parse rate 
(whether outputs follow the required action-reflection-score structure), 
and fit rate (full trajectory execution success with our complete 
reflection framework).

Without SFT, Qwen2-VL 2B shows noticeably lower semantic similarity 
and human evaluation scores, confirming that insufficient embodied 
prior leads to semantically unreliable reflections regardless of format. 
Qwen2.5-VL 3B and 7B without SFT already show high semantic quality 
but near-zero format parse rates, confirming that format is the only 
missing piece SFT addresses. After SFT, format is fixed across all 
models, but fit rate reveals the remaining story: Qwen2-VL 2B + SFT 
underperforms due to insufficient embodied prior; Qwen2.5-VL 3B + SFT 
hits the optimal fit rate; Qwen2.5-VL 7B + SFT degrades, consistent 
with stronger pretrained models being harder to adapt at test time.
\begin{table}[htb]
\centering
\small
\setlength{\tabcolsep}{4pt}
\begin{tabular}{lcccc}
\toprule
\textbf{Model} & \textbf{Sem.\ Sim.} & \textbf{Human Eval} & \textbf{Format Parse} & \textbf{Fit \%} \\
\midrule
Qwen2-VL 2B (no SFT)    & 0.59 & 76 & 0.19 & 4.7  \\
Qwen2.5-VL 3B (no SFT)  & 0.73 & 89 & 0.27 & 8.0  \\
Qwen2.5-VL 7B (no SFT)  & 0.75 & 91 & 0.31 & 11.9 \\
\midrule
Qwen2-VL 2B + SFT        & --- & --- & 0.95 & 48.5 \\
Qwen2.5-VL 3B + SFT      & --- & --- & 0.99 & 60.2 \\
Qwen2.5-VL 7B + SFT      & --- & --- & 0.99 & 52.7 \\
\bottomrule
\end{tabular}
\caption{Base model scale analysis. Without SFT, 3B/7B models have sufficient semantic quality but near-zero format parse rates; SFT fixes format. Qwen2.5-VL 3B hits the optimal fit rate; 7B degrades due to over-parameterization.}
\label{tab:model_scale}
\end{table}

\section{Retro-Reflection Quality Analysis}
\label{appendix:retro_quality}

% \subsection{Ablation: TTT on Immediate vs.\ Retrospective Reflection}

% \begin{table}[htb]
% \centering
% \small
% \setlength{\tabcolsep}{4pt}
% \begin{tabular}{lcccccc}
% \toprule
% \textbf{Method} & \textbf{Fit} & \textbf{Sel.} & \textbf{Prep.} & \textbf{Hybrid} & \textbf{Avg.} & \textbf{Overhead} \\
% \midrule
% TTT on Ext.\ Reflection  & 22.4 & 17.6 & 12.7 & 9.69 & 15.6 & $\sim$4$\times$ \\
% TTT on Retro-Reflection  & 44.7 & 32.4 & 31.7 & 25.8 & 33.7 & 1$\times$ \\
% \bottomrule
% \end{tabular}
% \caption{Retrospective reflection enables effective credit assignment at lower cost than immediate external reflection, which triggers at every step rather than at key milestones.}
% \label{tab:retro_ablation}
% \end{table}

In Table \ref{tab:retro_quality}, we analyze retro-reflection quality . 
\begin{table}[htbp]
\centering
\small
\setlength{\tabcolsep}{6pt}
\begin{tabular}{lcc}
\toprule
\textbf{Metric} & \textbf{BEHAVIOR} & \textbf{HM3D (OOD)} \\
\midrule
Revision Direction (\%)  & 99 & 95 \\
Human Eval --- Factual   & 98 & 96 \\
Human Eval --- Causal    & 95 & 96 \\
Human Eval --- Useful    & 99 & 95 \\
\bottomrule
\end{tabular}
\caption{Retro-reflection quality. Score revision directions align with ground-truth outcomes at $\geq$95\% even out-of-domain, and human evaluators rate reflections as highly factual, causally grounded, and useful.}
\label{tab:retro_quality}
\end{table}

\section{Search-Based Baseline Comparison}
\label{appendix:mcts}

We add Language-Guided MCTS (LG-MCTS), inspired by LLM-MCTS and MCTS-EP. LG-MCTS builds a search tree using $\pi_\theta$ for node expansion and $V_{\phi_i}$ as value estimator, with tree depth limited to 3 for tractability.

\begin{table}[htb]
\centering
\small
\setlength{\tabcolsep}{4pt}
\begin{tabular}{lcccccc}
\toprule
\textbf{Method} & \textbf{Fit} & \textbf{Sel.} & \textbf{Prep.} & \textbf{Hybrid} & \textbf{Avg.} & \textbf{Overhead} \\
\midrule
LG-MCTS          & 14.9 & 14.7 & 15.8 & 16.1 & 15.4 & $\sim$5--8$\times$ \\
Ours w/o TTT     & 6.4  & 11.8 & 19.0 & 12.9 & 12.5 & 0.5$\times$ \\
\textbf{Ours}    & 44.7 & 32.4 & 31.7 & 25.8 & 33.7 & 1$\times$ \\
\bottomrule
\end{tabular}
\caption{LG-MCTS is inferior to our full method at 5--8$\times$ cost. Search-based lookahead is ungrounded in physical uncertainties that can only be resolved through interaction; TTT, not search depth, is the key.}
\label{tab:mcts}
\end{table}

\section{Generalization to Human-Written Instructions}
\label{appendix:human_instructions}

We evaluate generalization beyond GPT-5-generated task descriptions, including deliberately ambiguous instructions and two completely held-out task types never seen during SFT.

\begin{table}[htb]
\centering
\small
\setlength{\tabcolsep}{4pt}
\begin{tabular}{lccccc}
\toprule
\textbf{Instruction Type} & \textbf{Fit} & \textbf{Sel.} & \textbf{Prep.} & \textbf{Hybrid} & \textbf{Avg.} \\
\midrule
GPT-Generated (current) & 44.7 & 32.4 & 31.7 & 25.8 & 33.7 \\
Human-Written           & 42.5 & 33.6 & 25.7 & 27.4 & 32.3 \\
\bottomrule
\end{tabular}
\caption{Performance generalizes to human-written instructions with marginal degradation.}
\label{tab:human_instructions}
\end{table}

\section{Commercial LLM Baselines}
\label{appendix:commercial_llms}

Table~\ref{tab:commercial} compares our framework against GPT-5, 
Gemini 2.5 Pro, and Qwen2.5 as zero-shot baselines on the Long-Horizon 
Household benchmark. All three receive only the scene configuration 
and task description at test time, the same partial observations as 
our model. Note that GPT-5's role in data generation is distinct: 
during initialization it has access to complete scene graphs including 
3D bounding boxes, whereas as a zero-shot baseline it operates under 
the same partial observability constraints.

Our framework outperforms all three commercial models by a large margin 
despite using a 3B base model, with GPT-5 achieving only 20.7\% and 
Gemini 2.5 Pro 19.7\% average success rate against our 33.7\%. This 
directly demonstrates that test-time learning, not model scale or 
general language model capability, drives the performance gains. 
Even the most capable frontier models cannot recover from failures 
within a single deployment episode without the reflection mechanisms 
our framework provides.
\begin{table}[htb]
\centering
\small
\setlength{\tabcolsep}{4pt}
\begin{tabular}{lccccc}
\toprule
\textbf{Method} & \textbf{Fit} & \textbf{Sel.} & \textbf{Prep.} & \textbf{Hybrid} & \textbf{Avg.} \\
\midrule
GPT-5            & 19.2 & 23.5 & 17.4 & 22.6 & 20.7 \\
Gemini 2.5 Pro   & 14.9 & 20.5 & 20.6 & 22.6 & 19.7 \\
Qwen2.5          & 12.8 & 14.6 & 14.3 & 15.1 & 14.2 \\
\midrule
\textbf{Ours}    & 44.7 & 32.4 & 31.7 & 25.8 & 33.7 \\
\bottomrule
\end{tabular}
\caption{Commercial LLM zero-shot baselines. Our framework outperforms all commercial models despite using a 3B base model, demonstrating that test-time learning, not model scale, drives the gains. GPT-5 as a zero-shot baseline receives only partial observations; its data-generation role uses complete scene graphs.}
\label{tab:commercial}
\end{table}

\section{GPT-5 Data Quality Validation}
\label{appendix:gpt5_quality}

Table~\ref{tab:gpt5_quality} validates the quality of GPT-5-generated 
initialization data along two axes: consistency with OmniGibson physics 
simulation (whether annotated outcomes match actual simulator dynamics) 
and consistency with human judgment (whether human annotators agree 
with GPT-5's trajectory assessments).

Raw GPT-5 scores achieve 85\% simulator consistency and 80\% human 
consistency. After executing all generated trajectories in OmniGibson 
and rejecting instances where annotated outcomes conflict with actual 
simulator dynamics, simulator consistency reaches 100\% by construction 
and human consistency rises to 98\%. This confirms that physics-simulation 
validation is an effective filter for ensuring initialization data 
quality, and that the resulting dataset provides a reliable foundation 
for the minimal SFT bootstrap described in Section~\ref{appendix:sft_analysis}.

\begin{table}[htb]
\centering
\small
\setlength{\tabcolsep}{6pt}
\begin{tabular}{lcc}
\toprule
\textbf{Method} & \textbf{Consistent w/ Sim.} & \textbf{Consistent w/ Human} \\
\midrule
GPT Raw Scores             & 85\% & 80\% \\
GPT + Sim.\ Validation     & 100\% & 98\% \\
\bottomrule
\end{tabular}
\caption{GPT-5 data quality after physics-simulation validation. Simulator filtering raises human consistency from 80\% to 98\%.}
\label{tab:gpt5_quality}
\end{table}

\begin{table}[htb]
\centering
\small
\setlength{\tabcolsep}{8pt}
\begin{tabular}{lc}
\toprule
\textbf{Dimension} & \textbf{Score} \\
\midrule
Task Naturalness     & 95 \\
Reflection Quality   & 99 \\
Score Calibration    & 97 \\
Diversity            & 92 \\
\bottomrule
\end{tabular}
\caption{Human evaluation of GPT-5-generated data quality.}
\label{tab:gpt5_human}
\end{table}

\section{Real-Robot Experiment Details}
\label{appendix:real_robot}

% \subsection{Setup}

We use a Franka Panda robotic arm in a cupboard fitting scenario. A camera mounted above the workspace captures RGB images from a top-down offset viewpoint matching the simulation configuration. High-level language commands are executed via two phases: (1) the robot moves to the object's registered position and grasps it with the parallel-jaw gripper; (2) it places the object at the target compartment location. After each action, the system provides binary success/failure feedback based on whether the object remains within target compartment bounds.

% \subsection{Quantitative Results}

% \begin{table}[htb]
% \centering
% \small
% \setlength{\tabcolsep}{6pt}
% \begin{tabular}{lcc}
% \toprule
% \textbf{Setting} & \textbf{Fit \%} & \textbf{Correct \%} \\
% \midrule
% Zero-shot, w/o RIA \& ROA          & 13.1 & 4.5  \\
% Fine-tuned on real, w/o RIA \& ROA & 20.7 & 7.2  \\
% Zero-shot, full                    & 40.6 & 13.4 \\
% Fine-tuned on real, full           & 44.2 & 16.6 \\
% \bottomrule
% \end{tabular}
% \caption{Real-robot results. The full model generalizes significantly better than ablations; real-world uncertainties (grasp imprecision, object slippage) make active failure recovery even more critical.}
% \label{tab:real_robot}
% \end{table}

\section{Hyperparameter Analyses for Cupboard Fitting Task}
\label{sec:hyperparameter_analysis}

\begin{figure*}[htbp]
\centering
\includegraphics[width=\linewidth]{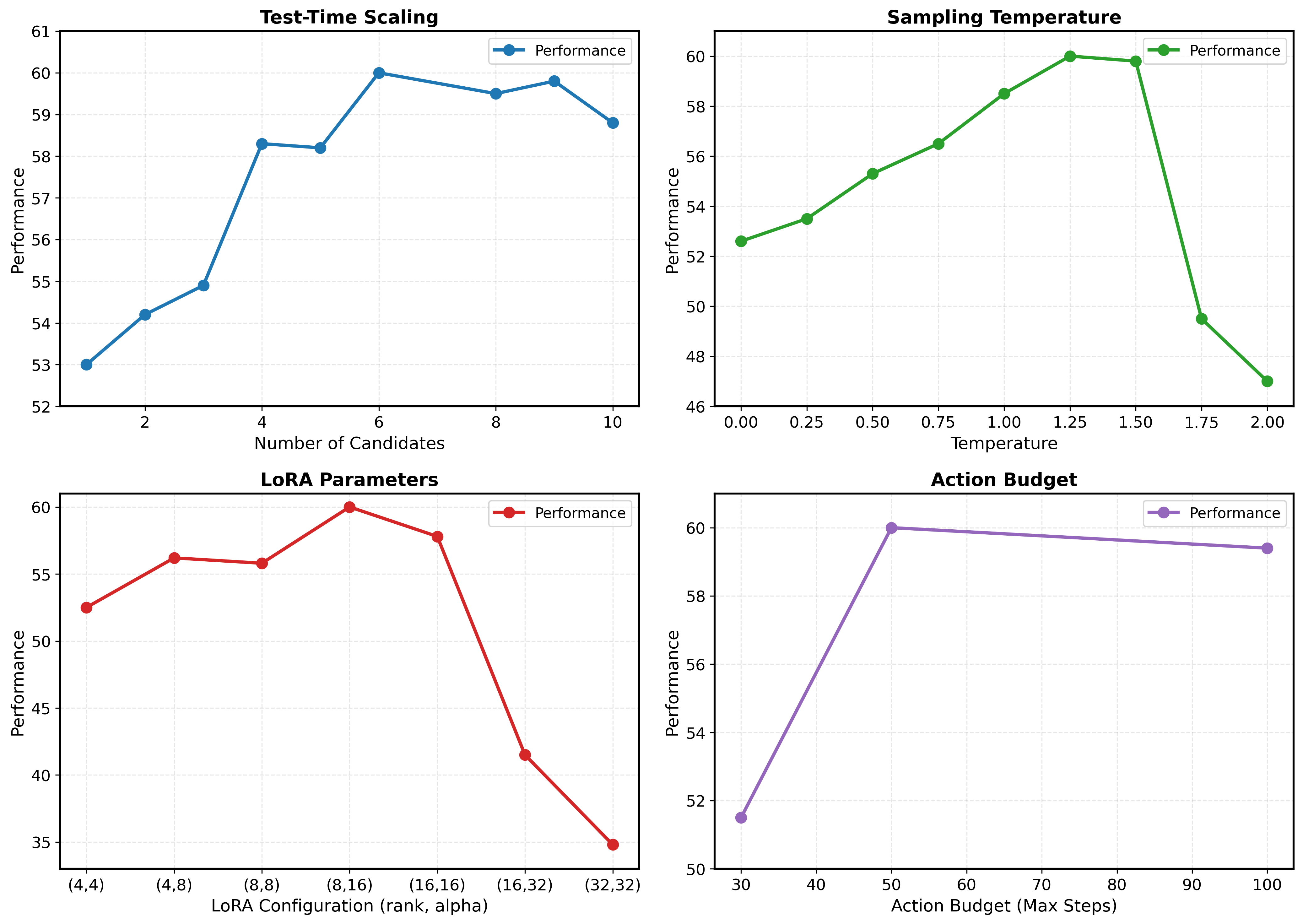}
\caption{\textbf{Hyperparameter ablation studies on Cupboard Fitting.} \textit{Top Left:} Performance vs. number of candidate actions. Peak performance (60.0\%) occurs at N=6 candidates, demonstrating that internal reflection effectively identifies superior actions from diverse pools. Beyond N=6, performance plateaus as excessive candidates add computational cost without improving the best candidate quality. \textit{Top Right:} Performance vs. sampling temperature. Optimal temperature range (T=1.25-1.5) balances candidate diversity with quality—temperatures below 0.5 produce overly similar candidates that limit reflection value, while temperatures above 1.75 generate incoherent actions that even accurate reflection cannot salvage. \textit{Bottom Left:} Performance vs. LoRA configuration (rank, alpha). The optimal configuration (r=8, $\alpha$=16) achieves 60.0\% performance, balancing adaptation capacity with training stability. Smaller configurations like (4,4) underfit with insufficient capacity (52.5\%), while larger configurations cause mode collapse during test-time training—(16,32) drops to 41.5\% and (32,32) collapses to 34.8\% as the model begins predicting identical outputs for all inputs, losing the ability to distinguish between different spatial configurations and task contexts. \textit{Bottom Right:} Performance vs. action budget (maximum steps). Performance improves dramatically from 30 steps (51.5\%) to 50 steps (60.0\%), but slightly degrades to 59.4\% at 100 steps, suggesting that excessive action budgets allow suboptimal exploration strategies that accumulate errors over longer horizons.}
\label{fig:cupboard_hyperparams}
\end{figure*}

We conduct ablation studies on four key hyperparameters that govern the reflection mechanism and show the results in Figure \ref{fig:cupboard_hyperparams}.

\subsection{Test-Time Scaling: Number of Candidate Actions}

The reflection-in-action mechanism generates $N$ candidate actions via sampling, scores each using the internal reflection model $V_{\phi_i}$, and executes the highest-scoring action. We vary $N \in \{1, 2, 3, 4, 5, 6, 8, 9, 10\}$ to measure the impact of candidate diversity on task performance (Figure~\ref{fig:cupboard_hyperparams}, top left).

\textbf{Results.} Performance improves monotonically from 53.0\% (N=1, greedy decoding) to 60.0\% (N=6), representing a 7 percentage point gain. This demonstrates that internal reflection scoring effectively identifies superior actions from diverse candidate pools. Beyond N=6, performance plateaus and slightly decreases to 58.8\% at N=10, suggesting diminishing returns from excessive candidates—likely due to increased computational cost without proportional quality improvements in the candidate pool.

\textbf{Analysis.} The optimal operating point at N=6 balances exploration breadth with computational efficiency. At N=1, the model lacks opportunities to reconsider suboptimal greedy choices. At N=2-5, expanding the candidate pool allows the internal reflection model to compare alternatives and avoid locally optimal but globally poor actions (e.g., placing a small object in a large compartment early). At N$\geq$6, the candidate pool may include too many low-quality options that add noise without improving the best candidate's quality, and the internal reflection model's scoring may become less reliable across excessively diverse samples.

\subsection{Sampling Temperature Analysis}

Temperature $T$ controls the sharpness of the probability distribution during action generation: low temperatures ($T \to 0$) produce near-greedy sampling, while high temperatures ($T \to \infty$) approach uniform sampling. We evaluate $T \in \{0.0, 0.25, 0.5, 0.75, 1.0, 1.25, 1.5, 1.75, 2.0\}$ (Figure~\ref{fig:cupboard_hyperparams}, top right).

\textbf{Results.} Performance exhibits a clear inverted-U relationship with temperature. At T=0.0 (deterministic greedy decoding), performance is 52.6\%.  Then it increases steadily through moderate temperatures, reaching a peak of 60.0\% at T=1.25, and maintains near-peak performance of 59.8\% at T=1.5. However, performance drops sharply at higher temperatures: 49.5\% at T=1.75 and 47.0\% at T=2.0.

\textbf{Analysis.} The optimal temperature range T=1.25-1.5 balances two competing factors: (1) \textit{Diversity for reflection}: Sufficient randomness generates meaningfully different candidates for the internal reflection model to compare, enabling it to identify strategic failures (e.g., blocking future placements) that greedy or near-greedy decoding would miss; (2) \textit{Quality preservation}: Excessive randomness (T$\geq$1.5) samples from the low-probability tail of the distribution, producing incoherent or physically infeasible actions that even accurate internal reflection cannot salvage. The sharp drop at T$\geq$1.75 indicates that overly stochastic sampling overwhelms the reflection mechanism's corrective capacity—no amount of scoring can recover from fundamentally poor candidate pools.

\subsection{LoRA Configuration Analysis}

For parameter-efficient test-time training, we apply Low-Rank Adaptation (LoRA) with varying rank $r$ and scaling factor $\alpha$. We evaluate configurations $(r, \alpha) \in \{(4, 4), (4, 8), (8, 8), (8, 16), (16, 16), (16, 32), (32, 32)\}$ to study the trade-off between adaptation capacity and training stability (Figure~\ref{fig:cupboard_hyperparams}, bottom left).

\textbf{Results.} Performance exhibits a sharp peak at the intermediate configuration. Small configurations show limited performance: (4,4) achieves 52.5\% and (4,8) reaches 56.2\%, indicating insufficient adaptation capacity. Performance peaks at (r=8, $\alpha$=16) with 60.0\%, our optimal configuration. However, larger configurations show dramatic performance degradation: (8,8) maintains 55.8\%, but (16,16) drops to 57.8\%, (16,32) plummets to 41.5\%, and (32,32) collapses catastrophically to 34.8\%—below even the no-adaptation baseline of 53.0\%.

\textbf{Analysis.} The LoRA rank controls the expressiveness of the adapter matrices, but excessive rank causes mode collapse during aggressive test-time training. At (r=4, $\alpha$=4) and (r=4, $\alpha$=8), the adapter capacity is too limited to capture nuanced spatial reasoning required for effective reflection—the model cannot adequately learn from retrospective feedback about blocking placements or strategic failures. The optimal configuration (r=8, $\alpha$=16) provides sufficient capacity to update internal reflection scoring and action selection policies based on task-specific feedback while maintaining stable optimization under high learning rates (0.2 for internal reflection, 0.01 for action model).

Larger configurations fail catastrophically due to mode collapse. At (r=16, $\alpha$=32) and beyond, the increased parameter space combined with aggressive learning rates and limited training data (10-15 retrospective examples per test-time training iteration) causes the model to collapse to predicting identical outputs for all inputs. Rather than learning task-specific spatial reasoning, the overparameterized adapters converge to degenerate solutions that ignore input variations. At (32,32), the model effectively stops distinguishing between different object configurations, compartment sizes, or task contexts—producing the same stereotyped action regardless of the actual scene state. This mode collapse is particularly severe because test-time training lacks the regularization and diverse data that prevent collapse in offline training.

\subsection{Action Budget Analysis}

We evaluate the impact of maximum action budget on task performance by varying the step limit from 30 to 100 steps (Figure~\ref{fig:cupboard_hyperparams}, bottom right).

\textbf{Results.} Performance shows a sharp initial gain followed by slight degradation. At 30 steps, performance is only 51.5\%, indicating insufficient budget to complete multi-object placement tasks. Performance peaks at 50 steps with 60.0\%, our optimal setting. However, extending the budget to 100 steps results in slight performance degradation to 59.4\%, a 0.6 percentage point drop.

\textbf{Analysis.} The low performance at 30 steps reflects task incompletion—agents frequently run out of steps before placing all objects, especially when early mistakes require exploration of alternative strategies. The 50-step budget provides sufficient runway for the agent to explore the environment, execute placements, recover from initial failures through reflection, and retry with corrected strategies learned via test-time training.

The counterintuitive degradation at 100 steps reveals a failure mode of excessive budgets: when given too many steps, agents exhibit suboptimal exploration patterns that accumulate errors over longer horizons. With a generous budget, the model may attempt more exploratory actions rather than committing to placements, leading to inefficient trajectories. Additionally, longer episodes provide more opportunities for compounding errors—a single poor placement early in a 100-step episode has more downstream consequences than in a tightly constrained 50-step episode where the agent must act decisively. This suggests that moderate action budgets not only improve computational efficiency but also serve as a useful inductive bias that encourages focused, goal-directed behavior in embodied agents.

\section{Single-Step Action Generation vs. Receding Horizon Planning}
\label{sec:receding}

A key design decision in our framework is single-step action generation: the agent generates and executes one action at a time, rather than planning action sequences through receding horizon control. We validate this choice through ablation experiments on the Cupboard Fitting benchmark.

\textbf{Experimental Setup.} We compare two variants of our full method: (1) \textit{Single-Step (Ours)}: generates one action, executes it, observes outcome, performs test-time training, then generates the next action; (2) \textit{Receding Horizon}: generates a complete action sequence (5-10 actions), executes only the first action, observes outcome, performs test-time training, then replans a new sequence from the updated state. Both variants use identical model architecture, training procedures, and test-time training mechanisms, differing only in planning granularity. The receding horizon approach requires approximately 5× more computation per step due to generating full sequences at each decision point.

\begin{table}[htbp]
\centering
\small
\caption{Single-step action generation vs. receding horizon planning on Cupboard Fitting. Despite using 5× more computation to plan sequences at each step, receding horizon shows degraded performance, demonstrating that single-step action generation is more effective for our reflective learning framework.}
\label{tab:planning_granularity}
\begin{tabular}{l|c|c|c}
\toprule
\textbf{Method} & \textbf{Fit} & \textbf{Correct } & \textbf{ Compute} \\
\midrule
Ours (Receding Horizon) & 57.8\% & 25.8\% & 5.0× \\
Ours (Single-Step) & 60.2\% & 25.3\% & 1.0× \\
\bottomrule
\end{tabular}
\end{table}

\textbf{Results.} Table~\ref{tab:planning_granularity} shows that receding horizon planning achieves only 57.8\% fit rate compared to 60.0\% for single-step action generation—a comparable performance but at the cost of 4x more computation per step. This demonstrates that planning full action sequences at each decision point not only increases computational cost but actually harms performance in our framework.

\textbf{Why Receding Horizon is Incompatible with Test-Time Training.} The performance gap reveals fundamental incompatibilities between sequence planning and reflective test-time training: 

(1) \textit{Wasted computation on unpredictable futures}: Receding horizon generates 5-action sequences at each step but executes only the first action, discarding 80\% of the computation. Critically, in our error-driven tasks, the outcomes of actions are inherently unpredictable before execution—whether an object fits in a compartment, whether placement will be stable, or whether grasping succeeds depends on precise physical interactions that cannot be reliably simulated. The model imagines 4 future actions based on assumed execution outcomes, but these assumptions are frequently wrong. When the first action fails or succeeds differently than predicted, the entire planned sequence becomes invalid. This wasted computation could instead generate more candidates for reflection-in-action (increasing N from 3 to 6) or perform additional test-time training epochs, both of which operate on actual execution outcomes rather than unreliable predictions.

(2) \textit{Learning from imagination conflicts with learning from reality}: Test-time training fundamentally relies on learning from actual execution feedback—the model updates its understanding of spatial constraints only after physically attempting placements and observing real outcomes (does the object fit? does it block other spaces?). However, generating 5-action sequences forces the model to predict these outcomes before they occur: "If I place object A here, then I can place object B there, then object C..." These predictions are made with the model's current (pre-update) understanding, which is precisely what test-time training aims to improve. The model must simultaneously optimize two conflicting objectives: (a) accurately predicting hypothetical future states for sequence generation, and (b) updating its beliefs based on actual execution outcomes that contradict those predictions. This creates optimization interference where gradients from test-time training (learned from reality) fight against the inductive bias from sequence generation (learned from imagination).

Our single-action approach eliminates this conflict by committing only to decisions that can be made based on current observations, executing the action to obtain ground truth feedback, updating the model with real outcomes, then making the next decision with improved understanding. This aligns perfectly with the test-time training paradigm: learn from actual experience, not imagined futures.

\textbf{Retrospective Reflection Provides Implicit Long-Horizon Planning.} A potential concern is that single-step action generation lacks the foresight that explicit sequence planning provides. However, retrospective reflection addresses this through a different mechanism. While receding horizon achieves long-horizon reasoning by explicitly generating future action sequences, our approach achieves it through learned anticipation: retrospective reflection re-evaluates past actions with hindsight about their long-term consequences, creating training signals that teach the internal reflection model to anticipate multi-step outcomes before execution. 

For example, when placing objects in a cupboard, an action that initially appears successful may be retrospectively downgraded when the agent discovers that this placement blocks the only space for a larger object. These retrospective scores train the internal reflection model to predict such long-horizon consequences at decision time—effectively distilling multi-step lookahead into single-step action evaluation. This learned implicit planning is more sample-efficient than explicit sequence generation: rather than exploring all possible future sequences at every step (most of which will be discarded), the agent learns which single actions lead to favorable long-term outcomes through accumulated experience.

\textbf{Computational Efficiency Analysis.} The 5× computational cost of receding horizon stems from generating full 5-action sequences at each decision step, only to execute the first action and discard the rest. This is particularly inefficient in our test-time training setting, where: (1) Sequence generation requires forward passes through the language model for all actions in the sequence; (2) The additional compute does not improve learning quality—test-time training still operates on single executed actions, so planning discarded sequences provides no learning benefit; (3) The saved computation in single-step action generation can be reallocated to improvements that actually help: generating more candidates (N=6 vs. N=3), performing more test-time training epochs, or using larger working memory windows for retrospective reflection.

Our results demonstrate that effective long-horizon reasoning in embodied agents need not come from explicit sequence planning. Instead, combining single-step action generation with retrospective reflection achieves superior performance at 5× lower computational cost by learning to anticipate long-term consequences rather than exhaustively simulating future possibilities.

\section{Experiments on Long-Horizon Household Tasks: More Details}
\label{app:behavior}

\subsection{Task Generation \& Validation Overview}
\label{behavior-task}
\textbf{Task Generation \& Execution.} We employ GPT-5 to generate task specifications through a carefully structured prompting procedure, adapting existing task templates  that emphasize long-horizon reasoning and failure recovery from the original BEHAVIOR-1K benchmark. 
% The generation prompt is detailed in Appendix~\ref{app:prompts}.
%consists of
% (1) scene context with complete room descriptions, object bounding boxes, and spatial relationships from the BEHAVIOR-1K scene graph; (2) a closed action vocabulary of navigation and manipulation primitives; (3) task design constraints requiring multi-room navigation and 3D spatial reasoning; (4) a curated object inventory from the BEHAVIOR Knowledge Base; (5) JSON format specifications; (6) reflection protocol instructions for both internal and external reflections; and (7) few-shot example tasks demonstrating valid task types.
Each generated task instance includes: (1) a task description, (2) 3-7 relevant rooms, (3) new objects with placement specifications, and (4) a complete trajectory with interleaved thoughts, actions, and reflections scores. Since we include scene graphs in the prompts to GPT-5, which provides object properties such as bounding boxes, GPT-5 can deduce potential failures (e.g., size mismatches) during generation.
% these predictions are subsequently verified through physics simulation in BEHAVIOR to ensure consistency with the actual scene dynamics. 
% Further implementation details are provided in Appendix~\ref{app:task_generation}.
% \textbf{Task Execution} 
We further execute the generated tasks in BEHAVIOR simulators to ensure consistency with the actual scene dynamics and provide ground-truth data for finetuning embodied LLMs.  We initialize the environment by loading the BEHAVIOR-1K scenes, placing new objects at designated locations, and positioning the robot at a default starting pose. At each step, the agent executes the given actions.
After every interaction, the system captures RGB-D observations, converted to point clouds which serve as the inputs to our Embodied LLMs. The simulator then performs physics-based verification and provides execution results that could be used for prompting external reflections. We also perform task verification checks by comparing the task execution results with GPT-5's generated task specifications.  Please refer to Appendix~\ref{app:behavior} for details about data generation, models and experiments.

\subsection{Why Build Upon BEHAVIOR?}

BEHAVIOR-1K~\cite{li2024behavior1k} provides an excellent foundation for embodied AI research with several key strengths: (1) \textbf{Photorealistic environments}: BEHAVIOR-1K features high-fidelity household scenes with realistic object models, physics simulation, and diverse room layouts across 1,000+ scenes; (2) \textbf{Rich object diversity}: The benchmark includes hundreds of object categories with varied sizes, shapes, and physical properties, enabling complex manipulation tasks; (3) \textbf{Standardized infrastructure}: BEHAVIOR-1K provides well-maintained simulation infrastructure, observation APIs, and action spaces that facilitate reproducible research.

However, BEHAVIOR-1K's original task design does not systematically stress two critical capabilities for reflective learning: (1) \textbf{Learning from failures}: Most BEHAVIOR-1K tasks are designed to be solvable with correct initial planning, without requiring agents to recover from or learn from execution failures. Tasks rarely include scenarios where early actions create downstream failures that only become apparent after multiple steps (e.g., placing a small object first that later blocks the only space for a larger object). (2) \textbf{Long-term dependencies}: The original benchmark emphasizes task completion but does not specifically design tasks around sequential dependencies where action order critically determines success, or where consequences of early actions remain hidden until much later in the episode.

To evaluate our reflective test-time training framework—which specifically learns from execution failures through retrospective reflection—we adapt BEHAVIOR-1K environments to create tasks that systematically incorporate these failure modes. We retain BEHAVIOR-1K's photorealistic scenes and simulation infrastructure while introducing task specifications that stress failure recovery, long-horizon credit assignment, and dependency reasoning. This allows us to leverage the strengths of BEHAVIOR-1K (realism, diversity, standardization) while evaluating capabilities (learning from failures, retrospective reasoning) that the original benchmark was not designed to measure.

\subsection{Task Categories}

We develop a systematic pipeline to generate Long-Horizon Household Tasks that stress error-driven adaptation in embodied agents. Our tasks are designed around four core failure modes common in real life: spatial reasoning errors, object selection mistakes, sequential dependency violations, and non-local planning failures.

\textbf{Task Categories and Failure Modes.}
We define four task categories, each targeting specific failure patterns:

\textit{Fitting Tasks} require agents to pack or place objects into constrained containers or surfaces. These tasks stress geometric reasoning and capacity constraints. Common failure modes include: (1) attempting to place oversized objects in small compartments, (2) blocking access to larger storage spaces with premature small-object placements, and (3) failing to recognize occlusion after placement. For example, placing a toy car in a box already containing a teddy bear may succeed physically but block future placements of larger items.

\textit{Selection Tasks} require agents to compare and retrieve items based on preferences or constraints. Failure modes include: (1) selecting suboptimal items when better alternatives exist in unexplored rooms, (2) committing to choices before exploring all options, and (3) failing to revise decisions when new information becomes available. A typical scenario involves retrieving vegetables for a meal where lettuce (preferred) is in one room and tomato (less preferred) is in another---agents that explore insufficiently may select the inferior option.

\textit{Preparation Tasks} involve sequential constraints and dependencies where actions must occur in specific orders. Common failures include: (1) attempting steps out of sequence (e.g., adding toppings before placing the base plate), (2) violating prerequisite conditions (e.g., trying to cook without retrieving ingredients first), and (3) missing intermediate setup steps. These tasks require agents to maintain action dependencies across multiple rooms and objects.

\textit{Hybrid Tasks} combine multiple failure modes within a single episode, requiring agents to simultaneously reason about spatial constraints, object preferences, and sequential dependencies across long horizons.

\begin{figure}[t]
    \centering
    \includegraphics[width=\linewidth]{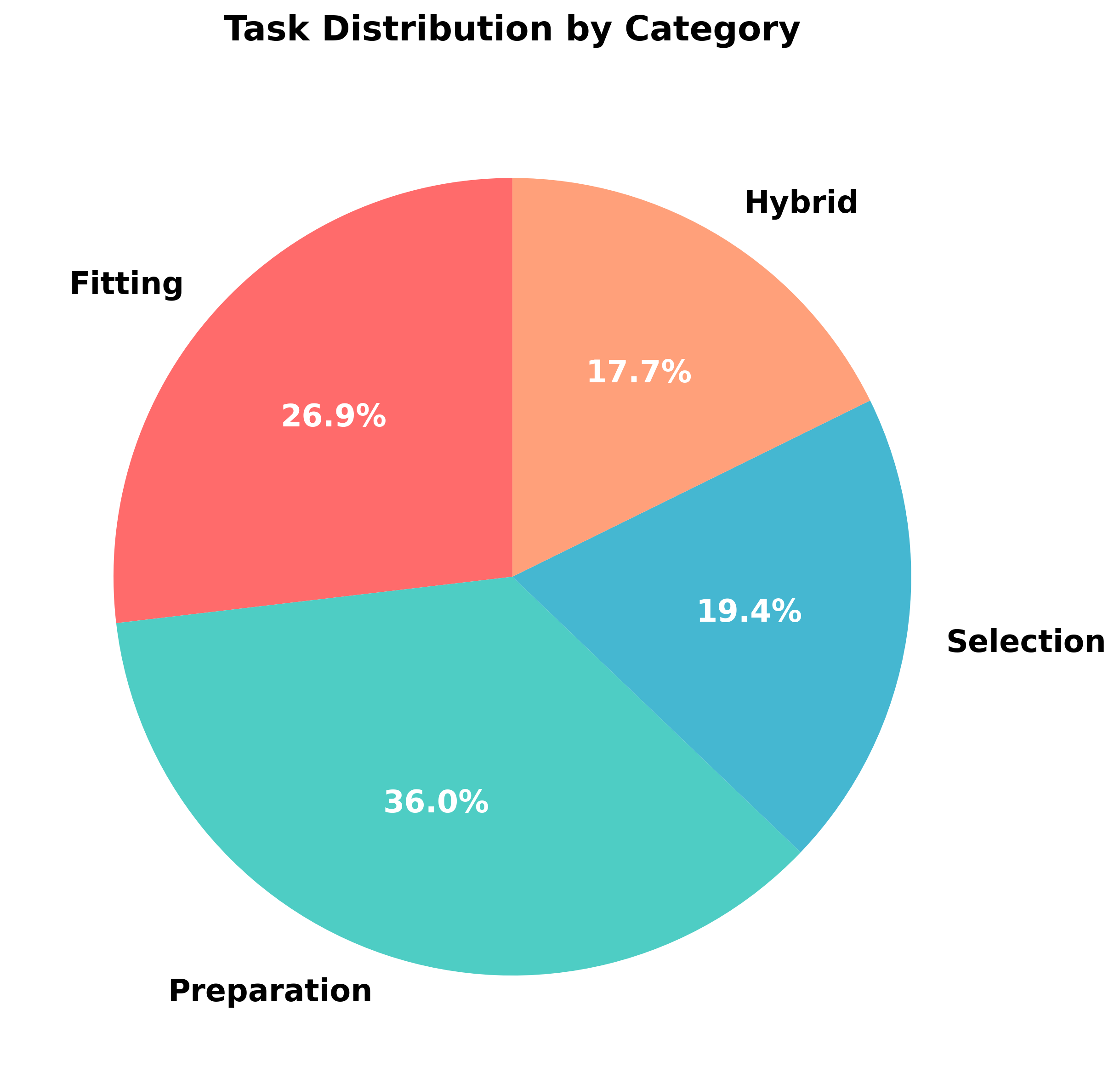}
    \caption{Distribution of task categories in the dataset. }
    \label{fig:task_distribution}
\end{figure}

The distribution of task categories can be found in Figure \ref{fig:task_distribution}.

\subsection{GPT-5 Task Generation Prompting Strategy}

We employ a carefully structured prompting approach to generate high-quality, physically plausible long-horizon tasks. Our prompting strategy consists of three key components: (1) comprehensive scene context provision, (2) explicit failure mode specification, and (3) structured output formatting with reflection annotations.

\textbf{Scene Context Provision.}
Each task generation prompt includes the complete scene graph from BEHAVIOR-1K, containing: room layouts with spatial relationships, existing objects in each room with 3D bounding boxes and affordance properties, navigable connections between rooms, and furniture placement. This rich context allows GPT-5 to reason about physical feasibility when proposing object placements and action sequences. For example, knowing that a table has dimensions 1.2m × 0.8m × 0.75m allows the model to avoid proposing placements of oversized objects.

\textbf{Failure Mode Specification.}
We explicitly instruct GPT-5 to incorporate specific failure scenarios into the generated trajectories. For Fitting tasks, we request scenarios where early placements block later optimal choices. For Selection tasks, we specify that preferred items should be in rooms requiring more exploration. For Preparation tasks, we request action sequences with complex dependencies where naive sequential execution fails. For each task category, we provide 2-3 concrete examples of desired failure patterns in few-shot demonstrations.

\textbf{Structured Output with Reflection Annotations.}
The prompt requires GPT-5 to generate not just action sequences, but complete trajectories with: (1) Each action wrapped in angle brackets (e.g., \texttt{<GO TO kitchen>}); (2) Internal reflections before each action with format \texttt{INTERNAL REFLECTION: [reasoning] | SCORE: [0-100]}; (3) Execution result annotations (\texttt{execution: success} or \texttt{execution: fail}); (4) External reflections after each executed action with format \texttt{PREVIOUS ACTION TO REFLECT ON: <action> | EXTERNAL REFLECTION: [assessment] | SCORE: [0-100]}; (5) Retrospective reflections at room exits with format \texttt{PREVIOUS ACTION TO REFLECT ON (retro): <action> | EXTERNAL REFLECTION: [hindsight assessment] | SCORE: [0-100]}.

Actions marked with \texttt{INTERNAL REFLECTION: ... | SCORE: [low score]} followed by the note ``score low. don't execute'' represent strategically poor choices that should be avoided through reflection-in-action. These actions may be physically feasible but lead to suboptimal task outcomes (e.g., picking up an inferior item when better alternatives exist). Yet they provide valuable data to train the internal reflection LLM (these actions are not used for training action LLM though). 

\textbf{Few-Shot Demonstrations.}
We provide 2-3 complete task examples for each task category, demonstrating the expected output format, reflection structure, and failure patterns. These examples show diverse scenarios: a Fitting task where placing small objects first blocks large object storage, a Selection task where exploring only nearby rooms leads to selecting inferior items, and a Preparation task where violating sequential dependencies causes failure.

% \textbf{Post-Generation Quality Checks.}
% After GPT-5 generates a task, we perform automated consistency checks before physical validation: (1) Verify all specified rooms exist in the scene graph; (2) Check that new object categories exist in the object inventory; (3) Validate that placement targets (furniture) are present in specified rooms; (4) Ensure action format compliance (proper angle brackets, valid action types); (5) Confirm presence of required reflection annotations. Tasks failing these checks are regenerated with additional constraints to address the specific issues.

\textbf{Prompt Iteration and Refinement.}
We iteratively refined the prompt through multiple rounds of generation and validation. Early versions produced tasks with: (1) Physically implausible object placements (e.g., large furniture items on small shelves); (2) Insufficient failure diversity (most tasks had similar error patterns); (3) Inconsistent reflection scores (high scores for objectively poor actions). We addressed these through: (1) Adding explicit bounding box information and size reasoning requirements; (2) Providing diverse few-shot examples spanning different failure modes; (3) Including calibration guidelines for score assignment (e.g., ``score 0-30 for actions that fail or lead to dead ends; 70-100 for optimal strategic choices'').

\textbf{Reflection Score Calibration.}
To ensure consistent score semantics across generated tasks, we provide GPT-5 with explicit calibration guidelines: Scores 0-20 indicate actions that fail physically or lead to immediate dead ends; scores 21-40 indicate poor strategic choices that succeed physically but create future problems; scores 41-60 indicate suboptimal but acceptable actions; scores 61-80 indicate good strategic choices; scores 81-100 indicate optimal or near-optimal actions. This calibration ensures that scores are meaningful training signals for the reflection models rather than arbitrary numbers.

% \textbf{Automated Task Generation with GPT-5.}
% We employ GPT-5 to generate task specifications through structured prompting. Each generation receives: (1) the target BEHAVIOR-1K scene with complete scene graph including object bounding boxes and room layouts, (2) the task category (Fitting/Selection/Preparation/Hybrid), (3) constraints on number of rooms (3-7) and objects (3-5), and (4) expected failure scenarios to include.

% The generator produces task specifications containing: (a) \textit{Task description}: Natural language instruction including user preferences where applicable; (b) \textit{New objects}: List of objects to be added with categories, placement locations, and spatial relationships; (c) \textit{Target rooms}: Subset of scene rooms relevant to the task; (d) \textit{Ground-truth trajectory}: Complete action sequence with internal reflections, execution results, and external reflections.

% Crucially, the trajectory includes \textit{expected failure annotations}. Each action is labeled with anticipated execution results (\texttt{success}/\texttt{fail}) and reflection scores (0-100). Actions marked with ``score low; don't execute'' represent poor strategic choices that should be avoided through reflection-in-action, while actions with ``execution: fail'' represent physically infeasible actions that reveal environmental constraints only after execution.

\subsection{Physical Validation in BEHAVIOR.}
Raw GPT-5 outputs may contain physically implausible scenarios or inconsistent spatial reasoning. We validate each generated task through execution in BEHAVIOR OmniGibson physics simulation:

\textit{Object Placement Validation}: We verify all new objects can be physically placed at specified locations using sampling-based kinematics (\texttt{sample\_kinematics}). Objects that fail placement (due to size mismatches, collision constraints, or stability issues) trigger task rejection. We use uniform scaling for all objects to avoid non-orthogonal transform errors.

\textit{Trajectory Execution Verification}: We execute the ground-truth trajectory action-by-action, verifying that: (1) Navigation actions (\texttt{GO TO}) successfully place the robot in target rooms; (2) Manipulation actions (\texttt{PICK UP}, \texttt{PUT DOWN}) complete as annotated; (3) Expected failures actually fail (confirming physical constraints match annotations); (4) Action sequences respect object affordances and spatial constraints. Tasks where any expected-success action fails, or any expected-failure action succeeds, are rejected as inconsistent. This ensures our evaluation measures genuine agent learning rather than dataset annotation errors.

\textit{Observation Generation}: During validation, we generate 3D point cloud observations for each room after every interaction step (navigation, pickup, placement). We capture observations from three external camera viewpoints positioned around the robot at fixed relative poses. For each viewpoint, we: (1) Capture RGB-D images at 560×560 resolution; (2) Convert depth to point clouds using camera intrinsics; (3) Transform point clouds from camera frame to robot base frame; (4) Store point clouds (stacked across viewpoints) as \texttt{.npy} files; (5) Save corresponding RGB images as \texttt{.png} files. Each room observation is stored in a unique directory named \texttt{\{room\_name\}\_\{step\_idx\}}, allowing models to access the most recent observation per room at any point during inference.

\subsection{Training Data Construction}

From validated trajectories, we extract training data for three model components. All training examples include point cloud observations stored as file paths, which are loaded and processed by the 3D vision encoder during training.

\textbf{Action Training Data.}
For each step $t$ in a validated trajectory, we create training examples of the form:
\begin{align*}
\text{Input: } & \{\text{Task}, \text{All Rooms}, \text{Explored Rooms}, \\
& \text{Current Room}, \text{Observations}, \\
& \text{Previous Action}, \text{Previous External Reflection}\} \\
\text{Output: } & \text{Action}_t
\end{align*}
Observations consist of point cloud paths for each explored room (most recent per room). 

\textbf{Internal Reflection Training Data.}
For each action (including non-executed low-score actions), we create examples:
\begin{align*}
\text{Input: } & \{\text{Task}, \text{All Rooms}, \text{Explored Rooms}, \\
& \text{Current Room}, \text{Observations}, \\
& \text{Previous Action}, \text{Previous External Reflection}\} \\
& \text{Potential Action}_t\} \\
\text{Output: } & \{\text{Internal Reflection}_t, \text{Score}_t\}
\end{align*}
This includes actions marked ``don't execute''---the model must learn to score these actions low during internal reflection to prevent execution.

\textbf{External Reflection Training Data.}
After each executed action, we create examples:
\begin{align*}
\text{Input: } & \{\text{Task}, \text{All Rooms}, \text{Explored Rooms}, \\
& \text{Current Room}, \text{Observations (before and after)}, \\
& \text{Previous Action}, \text{Previous External Reflection}\} \\
& \text{Executed Action}_t, \text{Execution Result}_t\} \\
\text{Output: } & \{\text{External Reflection}_t, \text{Score}_t\}
\end{align*}
Execution results indicate \texttt{success} or \texttt{fail}, and observations include both pre-action and post-action point clouds to enable change detection.

\textbf{Retrospective Reflection Training Data.}
For retrospective reflection, we identify room transitions (marked by \texttt{EXIT} actions) and collect all actions taken in that room. For each historical action $a_j$, we create:
\begin{align*}
\text{Input: } & \{\text{Task}, \text{Context}, \text{Current Observations}, \\
& \text{Action}_j, \text{Room Action History}, \\
& \text{Last Reflection}_j\} \\
\text{Output: } & \{\text{Retro Reflection}_j, \text{Updated Score}_j\}
\end{align*}
The prompt includes all actions and their external reflections from the current room window, allowing the model to re-evaluate $a_j$ with hindsight about downstream consequences.

\subsection{Model Architecture and Training}
\label{behavior-details}
\textbf{Base Architecture.}
We build our 3D vision-language-action model on LLaVA-3D~\cite{zhu2024llava3d}, which processes point clouds through a 3D encoder and fuses them with language instructions via a multimodal projector. The architecture consists of: (1) A 3D point cloud encoder that extracts spatial features; (2) A vision-language projector that aligns 3D features with language embeddings; (3) A Llama-based language model backbone (7B parameters) for reasoning and generation.

\textbf{Unified Multi-task Pretraining.}
Rather than training three separate models, we first train a single unified model on all three reflection modes (action generation, internal reflection, external reflection) using task-specific prompts to distinguish modes. This enables cross-mode knowledge transfer: the model learns shared representations for spatial reasoning and object affordances that benefit all three capabilities. We fine-tune LLaVA-3D-7B with learning rate $2 \times 10^{-5}$, batch size 8 per GPU, gradient accumulation over 4 steps, for 3 epochs using AdamW optimizer with weight decay 0.01. Following prior work~\cite{hu20253dllmmemlongtermspatialtemporalmemory}, we maintain fused point cloud observations from previous steps to provide the model with spatial memory across the episode.

\textbf{Test-Time Model Instantiation.}
At deployment, we instantiate three separate copies of the finetuned unified model: $\pi_\theta$ (action generation model), $V_{\phi_i}$ (internal reflection model, updated via test-time training), and $V_{\phi_e}$ (external reflection model, frozen). This architecture allows selective adaptation: only $V_{\phi_i}$ and $\pi_\theta$ update their parameters during test-time deployment, while $V_{\phi_e}$ remains fixed to provide stable assessment signals.

\subsection{Test-Time Training Configuration}

\textbf{LoRA Configuration.}
For memory-efficient test-time training, we apply LoRA with rank $r = 4$, alpha $\alpha = 8$, dropout rate 0.15, targeting \texttt{q\_proj} and \texttt{v\_proj} modules.

\textbf{Internal Reflection Model Training.}
We train $V_{\phi_i}$ via supervised learning to predict retrospective reflections  with learning rate $5 \times 10^{-5}$, 3 epochs, gradient clipping at 0.3, and negative log-likelihood loss over reflection text.

\textbf{Action Model Training via REINFORCE.}
We train $\pi_\theta$ via policy gradient using retrospective scores as rewards: SGD optimizer with learning rate $1 \times 10^{-3}$, 3 epochs, gradient clipping at 0.5, and reward transformation $r = 2(s_r/100) - 1$ mapping [0,100] scores to [-1,1] rewards.

\textbf{Regularization Strategy.}
To prevent catastrophic forgetting during test-time training, we include regularization examples in $\mathcal{D}_{\text{train}}$: we sample unexplored actions from the action space and use the current internal reflection model's outputs as targets, with a 50\% retrospective examples and 50\% regularization examples mixing ratio. This anchors the model to its pretrained knowledge for regions of the state space not covered by recent experience.

\subsection{Evaluation Protocol and Baselines}
\label{baselines}
\textbf{Train-Test Split.}
We ensure zero overlap between finetuning and evaluation: (1) No shared task descriptions---evaluation tasks have completely different natural language instructions; (2) No shared scenes---different scene instances from BEHAVIOR-1K; (3) No shared object placements---all object configurations are unique; (4) No shared trajectories---action sequences and reflection patterns differ. This tests generalization to novel tasks rather than memorization.

\textbf{Success Criteria.}
A task succeeds if and only if: (1) All required objects reach target locations; (2) All spatial constraints are satisfied (e.g., inside, on top of); (3) All preference constraints are met (e.g., selecting preferred items); (4) All sequential dependencies are respected; (5) Task completion occurs within the action budget (30 steps).

\textbf{Deployment Procedure.}
For each test task, we: (1) Initialize agent in the first room with task description; (2) Execute action generation + internal reflection + external reflection loop; (3) Trigger retrospective reflection at room exits or after $K=5$ steps; (4) Perform test-time training when retrospective reflections accumulate; (5) Continue until task completion or action budget exhaustion. The agent receives no human feedback during deployment---all learning signals come from self-generated reflections.

\textbf{Baseline Implementations.}
We implement the following baselines: \textit{Reflexion}~\cite{shinn2023reflexion} maintains a text buffer of past reflections and includes them in prompts for future actions, generating verbal critiques after each step without parameter updates. \textit{Self-Refine}~\cite{madaan2023selfrefine} iteratively improves actions through self-critique and revision cycles, allowing up to 3 refinement iterations per action before execution. We revise the above two baselines to incorporate multimodal inputs. \textit{ReflectVLM} adapts the reflection mechanism from~\cite{feng2025reflectiveplanningvisionlanguagemodels} using learned value functions for action scoring, with a separate value head trained on our training data. \textit{3DLLM-Mem}~\cite{hu20253dllmmemlongtermspatialtemporalmemory} maintains fused point cloud observations from all previous steps and previous-step execution results as context, providing spatial memory without explicit reflection or test-time training. \textit{PPO}~\cite{schulman2017proximalpolicyoptimizationalgorithms} and \textit{DreamerV3}~\cite{zhou2024robodreamerlearningcompositionalworld} are trained as reinforcement learning baselines: PPO uses on-policy policy gradient with clipped surrogate objective and GAE for advantage estimation, while DreamerV3 learns a world model from observations and trains a policy in the learned latent space. Both RL baselines are trained for the same total number of environment interactions as our supervised fine-tuning phase to ensure fair compute comparison. All baselines use the same LLaVA-3D-7B backbone with identical finetuning procedures where applicable for fair comparison.

\section{Cupboard Fitting: More Details}
\label{appendix:cupboard_details}

\subsection{Cupboard Setup}
\label{cupboard_setup}
The task environment consists of three key components. First, a multi-compartment cupboard structure with 6-8 compartments of varying sizes and colors, each defined by precise 3D bounding boxes. Second, a set of 6-10 colored geometric objects that must be placed into compartments. Third, a Franka Panda robotic arm controlled through high-level natural language commands such as ``pick up the red apple" or ``put the blue object in the green compartment."  The agent's objective is to place all objects into the cupboard such that each fits completely within compartment boundaries, while multiple objects may share compartments when space permits. Each task has only one valid solution where all objects fit within their designated compartments. Success requires reasoning about object-compartment compatibility, multi-object spatial packing, and long-horizon dependencies where early placement decisions affect later possibilities. After each action, the environment provides execution status through forward simulation, and updated visual observations. We define two evaluation metrics: correct rate measures the percentage of objects placed in their correct target compartments, while fit rate measures the percentage of objects successfully placed in any compartment.

\subsection{Base Model and Supervised Fine-tuning.}
We build upon Qwen2.5-VL-3B as our base vision-language model for the Cupboard Fitting benchmark. We employ unified multi-task supervised fine-tuning on action generation, internal reflection, and external reflection tasks using: global batch size 128, learning rate $1 \times 10^{-5}$, weight decay 0.1, trained for 3 epochs using AdamW optimizer. Task-specific prompts distinguish between the three reflection modes, enabling cross-task knowledge transfer.

\subsection{Test-Time Training Variants.}
We implement two test-time training variants to study the trade-off between adaptation capacity and computational efficiency:

\textit{Base-Weight Test-Time Training for Reflection-on-Action:} This variant updates all non-visual parameters during deployment. We freeze visual encoder parameters (identified by \texttt{visual} or \texttt{vision} in parameter names) but update all language model parameters. For the action model trained via REINFORCE, we use SGD optimizer with learning rate $1 \times 10^{-3}$, weight decay $1 \times 10^{-4}$, zero momentum, training for 3 epochs. For the internal reflection model trained via supervised learning, we use SGD optimizer with learning rate $5 \times 10^{-5}$, weight decay $1 \times 10^{-4}$, zero momentum, training for 3 epochs. This approach provides maximum adaptation capacity but requires updating millions of parameters.

\textit{LoRA Test-Time Training for Reflection-on-Action:} For memory-efficient adaptation, we apply Low-Rank Adaptation with rank $r = 8$, alpha $\alpha = 16$, dropout rate 0.1, targeting all linear layers except \texttt{lm\_head}, \texttt{embed\_tokens}, and visual encoder components. The LoRA adapters are applied to both internal reflection and action models using the PEFT library. For the action model, we use SGD with learning rate $1 \times 10^{-2}$, weight decay $1 \times 10^{-4}$, zero momentum, trained for 3 epochs. For the internal reflection model, we use SGD with learning rate $0.2$, weight decay $1 \times 10^{-4}$, zero momentum, trained for 3 epochs. The higher learning rates compensate for the reduced parameter count—LoRA updates only the low-rank adapter parameters, reducing trainable parameters by over 95\% while maintaining comparable performance.

\section{More Qualitative Examples}

\begin{figure*}[htbp]
    \centering
    \includegraphics[width=0.82\linewidth]{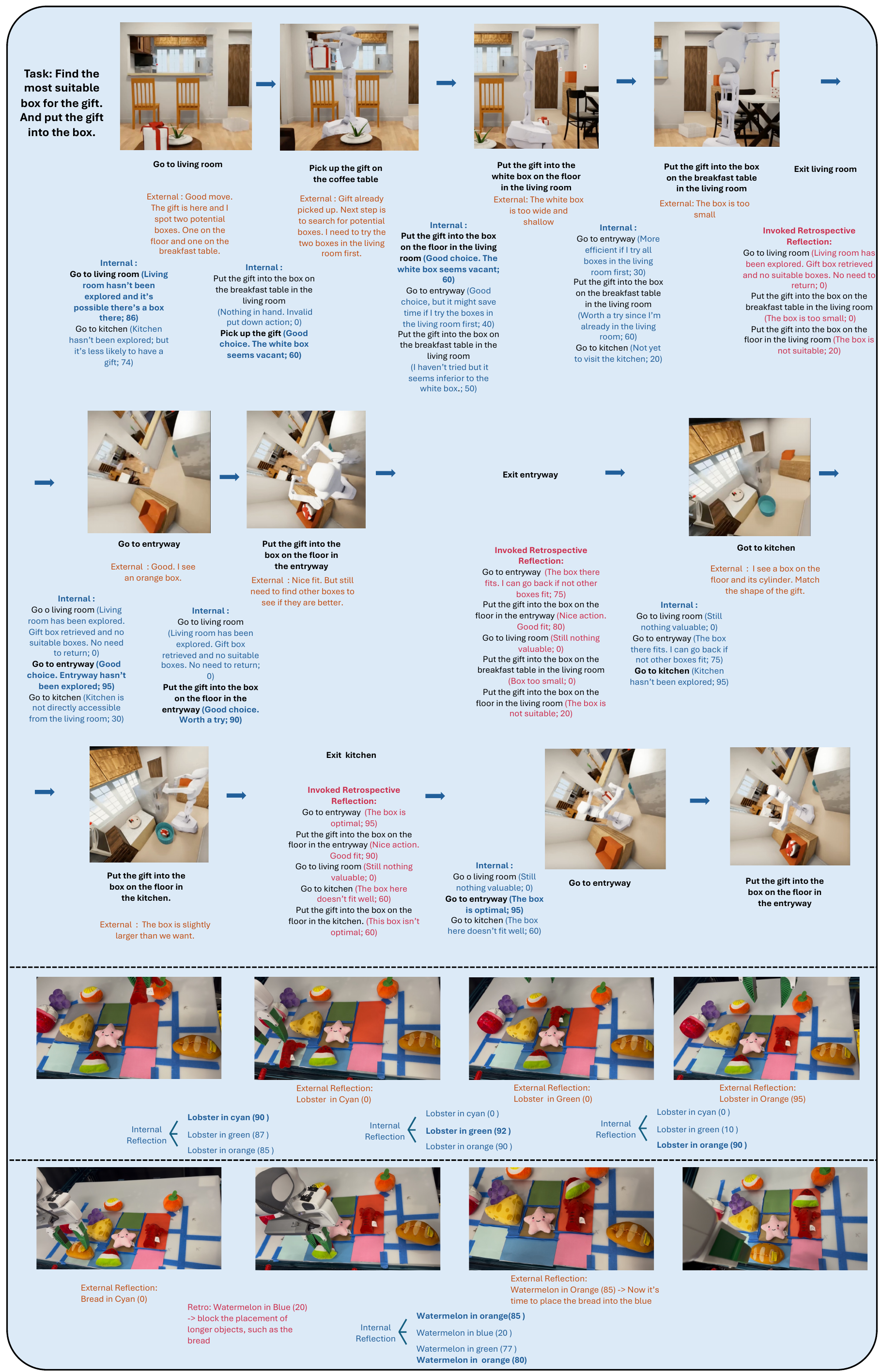}
    \caption{Additional qualitative examples. Blue text: internal reflection for candidate selection. Orange text: external reflection after execution. Red text: retrospective reflection and model updates. Reflection scores shown in brackets. Steps and reflections simplified for presentation.}
    \label{fig:appendix_qualitative}
\end{figure*}

In Figure~\ref{fig:appendix_qualitative}, we show additional qualitative examples demonstrating how the model improves over the course of a deployment episode through reflection mechanisms.

\end{document}